\documentclass[10pt,twocolumn,letterpaper]{article}

\usepackage{cvpr}              

\usepackage{bm}
\definecolor{cvprblue}{rgb}{0.21,0.49,0.74}
\usepackage[pagebackref,breaklinks,colorlinks,allcolors=cvprblue]{hyperref}

\def\paperID{****} 
\def\confName{CVPR}
\def\confYear{2026}

\title{ProGIC: Progressive and Lightweight Generative Image Compression with Residual Vector Quantization}

\author{Hao Cao$^{1}$ \quad Chengbin Liang$^{1}$ \quad Wenqi Guo$^{1,*}$ \quad Zhijin Qin$^{1,2,*}$ \quad Jungong Han$^{1}$\\
$^{1}$Tsinghua University \quad
$^{2}$State Key Laboratory of Space Network and Communications\\
{\tt\small \{caoh24, lcb25\}@mails.tsinghua.edu.cn} \quad {\tt\small wenqiguo@mail.tsinghua.edu.cn} \\ {\tt\small \{qinzhijin, jghan\}@tsinghua.edu.cn}
}

\begin{document}
\maketitle

\begingroup
\renewcommand\thefootnote{}
\footnotetext{* Corresponding authors.}
\endgroup

\begin{abstract}

Recent advances in generative image compression (GIC) have delivered remarkable improvements in perceptual quality. However, many GICs rely on large-scale and rigid models, which severely constrain their utility for flexible transmission and practical deployment in low-bitrate scenarios. To address these issues, we propose \textbf{Pro}gressive \textbf{G}enerative \textbf{I}mage \textbf{C}ompression (ProGIC), a compact codec built on residual vector quantization (RVQ). In RVQ, a sequence of vector quantizers encodes the residuals stage by stage, each with its own codebook. The resulting codewords sum to a coarse-to-fine reconstruction and a progressive bitstream, enabling previews from partial data. We pair this with a lightweight backbone based on depthwise-separable convolutions and small attention blocks, enabling practical deployment on both GPUs and CPU-only devices. Experimental results show that ProGIC attains comparable compression performance compared with previous methods. It achieves bitrate savings of up to 57.57\% on DISTS and 58.83\% on LPIPS compared to MS-ILLM on the Kodak dataset. Beyond perceptual quality, ProGIC enables progressive transmission for flexibility, and also delivers over $10\times$ faster encoding and decoding compared with MS-ILLM on GPUs for efficiency.

\end{abstract}    
\section{Introduction}
Recent progress in image compression has made it possible to achieve better visual quality at higher compression ratio. However, traditional codecs~\cite{jpeg, vtm} and learned image codecs~\cite{lic-balle2018variational, hpcm} (optimized for MSE) often result in blurry images and blocking artifacts at low bitrates. This reveals a gap between image perceptual quality and pixel-level quality~\cite{perception}. To mitigate this issue, generative image compression (GIC) has emerged, aiming to improve perceptual quality by synthesizing reasonable details. Pioneering works~\cite{hific, ms-illm} used generative adversarial networks (GANs)~\cite{gan} to create missing image details. Other studies~\cite{vq-gic, vq-mao2024extreme, fu2023vector, vq-DLF} explored vector quantization (VQ)~\cite{vq-vae} to map low-quality features to better ones from a pretrained codebook. Building upon diffusion models~\cite{ddpm}, recent methods~\cite{diffeic,diffusion-StableCodec,diffusion-xue2025one,oscar} enhanced detail generation through pretrained Stable Diffusion~\cite{stable-diffusion}.

\begin{figure}[t]
  \centering
  \includegraphics[width=0.95\linewidth]{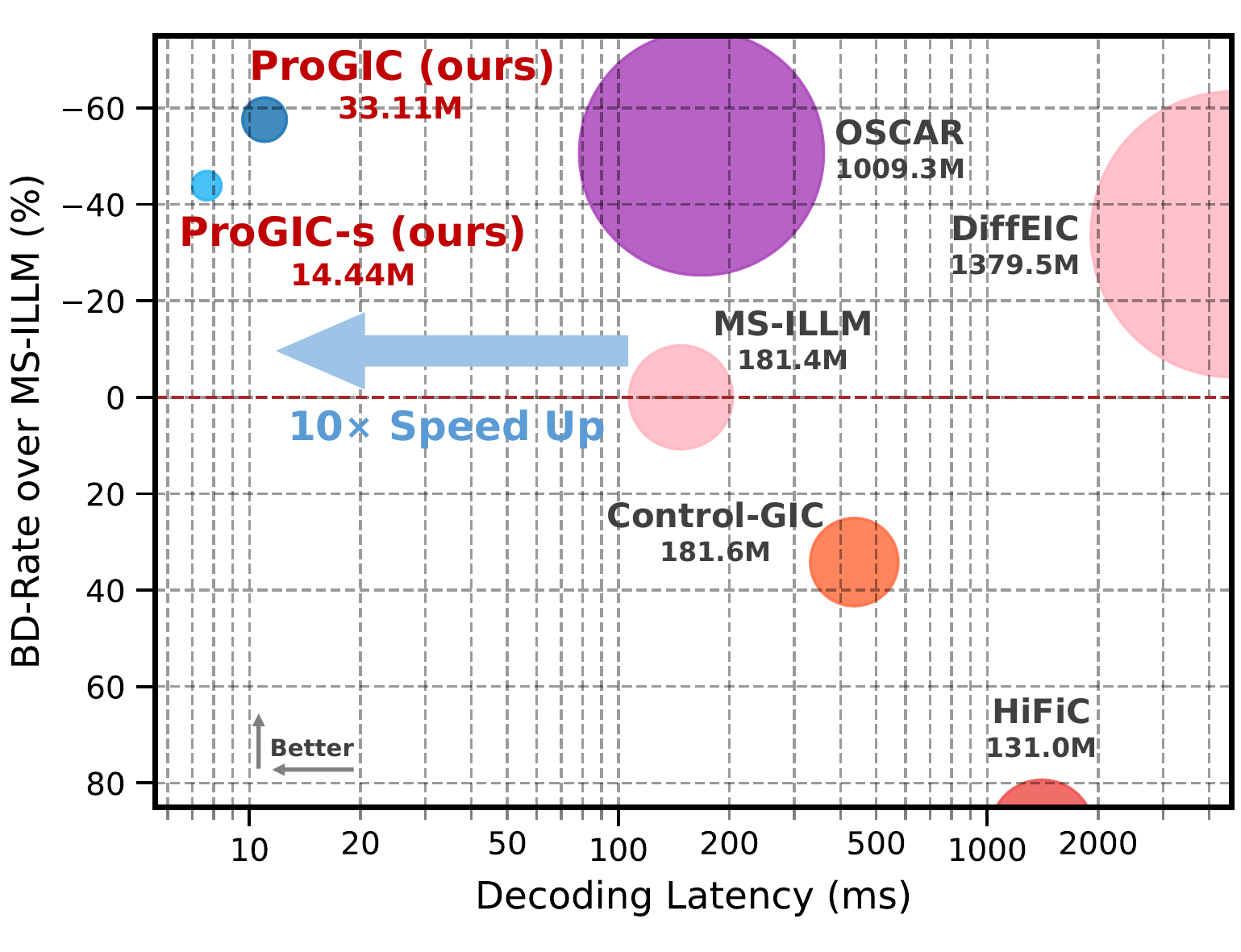}
  \caption{BD-rate vs. Decoding Latency on the Kodak dataset measured with DISTS on one NVIDIA A100 GPU. The proposed ProGIC attains competitive BD-rate while substantially reducing decoding latency. Upper-left indicates better.}
  \label{fig:BD-rate-fig}
\end{figure}

\begin{figure*}[t]
  \centering
  \includegraphics[width=0.95\linewidth]{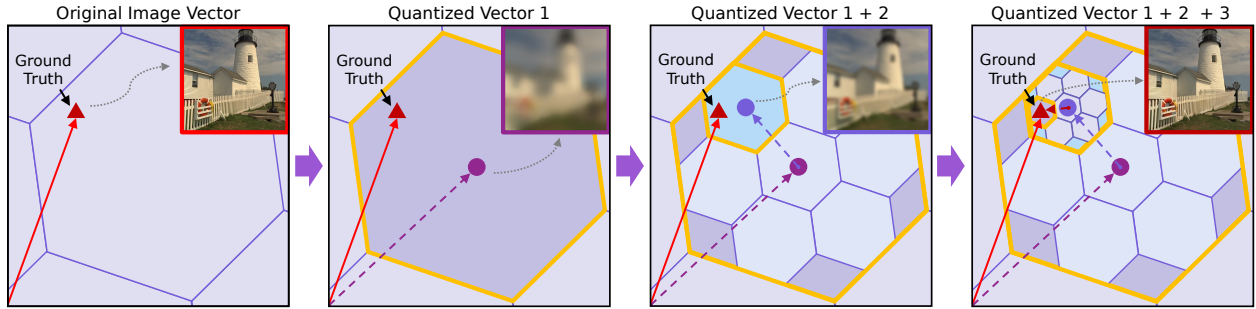}
   \caption{Conceptual illustration of the motivation behind ProGIC. The original image vector is approximated by a base vector plus a sequence of residual vectors, yielding progressively improved reconstructions.}
  \label{fig:prog_illustrate}
\end{figure*}

However, GICs face two main challenges. First, they are usually designed for extremely low-bitrate scenarios~\cite{vq-gic,vq-DLF}, where bandwidth is scarce and only an initial fragment of the bitstream can be sent. Under such constraints, transmitting the entire image is time-consuming, while practical needs is a rapid preview within a short time~\cite{satellite}. Progressive decoding addresses this need by turning partial streams into usable intermediate reconstructions. Nevertheless, existing methods still require the complete bitstream before producing a viable result. Although a few non-generative codecs support progressive decoding~\cite{lic-progdtd,lic-ctc,lic-dpict,diffusion-prog}, work specifically on progressive GIC remains limited.

Second, such low-bitrate scenarios often deploy on edge devices with constrained computational resources~\cite{satellite}. However, many works~\cite{oscar, diffusion-StableCodec, diffeic, llm-gao2025exploring} rely on large models with high parameter counts and slow inference, as shown in \cref{fig:BD-rate-fig}. These two challenges raise a critical question: \textit{How can we build flexible generative image compression for both bandwidth- and compute-limited environments?}

To address this question, we propose \textbf{Pro}gressive \textbf{G}enerative \textbf{I}mage \textbf{C}ompression (ProGIC), an image codec designed for limited-bandwidth and limited-compute settings. Some Prior works~\cite{vq-mao2024extreme, cgic} employed a single codebook for compression, which limited representation capacity. Inspired by residual vector quantization (RVQ) in discrete speech representation~\cite{speech-soundstream, speech-dac}, we model the image latent as a sum of quantized residual vectors from multiple codebooks. As shown in \cref{fig:prog_illustrate}, ProGIC refines the image latent in stages. Each stage quantizes the residual of the current reconstruction, and summing outputs in all stages yields a finer latent. This enables early previews without waiting for the full bitstream. Inspired by the success of lightweight models~\cite{dcvc-rt}, we introduce a lightweight backbone based on depthwise-separable convolutions with small attention~\cite{lic-attention} to further improve efficiency, enabling deployment on GPU and CPU-only mobile devices.

With this design, ProGIC achieves comparable compression performance across multiple datasets and metrics. \cref{fig:BD-rate-fig} summarizes the trade-off between performance and decoding latency, where ProGIC offers favorable performance and speed. On the Kodak dataset~\cite{Kodak}, ProGIC yields BD-rate~\cite{bd-rate} savings of 57.57\% on DISTS~\cite{dists} and 58.83\% on LPIPS~\cite{lpips} compared to MS-ILLM~\cite{ms-illm}, and it also surpasses recent diffusion-based methods~\cite{diffeic, oscar}. ProGIC delivers substantial runtime gains, achieving over $10\times$ faster encoding and decoding than MS-ILLM. We also provide a real-world use case on CPU-only mobile devices in the supplementary material, demonstrating the practicality of ProGIC.

Our main contributions are summarized as follows:

\begin{itemize}
    \item We propose ProGIC, which represents the latent by decomposing the quantization error into multiple residual stages. It supports coarse-to-fine reconstruction from partial bitstreams, enabling rapid previews under extreme bandwidth constraints.
    \item By combining RVQ with a lightweight backbone, we propose a compact generative image codec that enables real-world deployment for resource-limited devices.
    \item Experiments show ProGIC achieves comparable compression performance while offering notable speedups on both GPU and CPU-only mobile devices.
\end{itemize}

\section{Related Works}
\begin{figure*}[t]
  \centering
  \begin{subfigure}[c]{0.76\linewidth}
    \includegraphics[width=0.9999\linewidth]{fig/frame_work.pdf}
    \caption{Residual Vector Quantization based GIC.}
    \label{fig:short-a}
  \end{subfigure}
  \hfill
  \begin{subfigure}[c]{0.23\linewidth}
    \begin{subfigure}{0.9999\linewidth}
      \includegraphics[width=0.9999\linewidth]{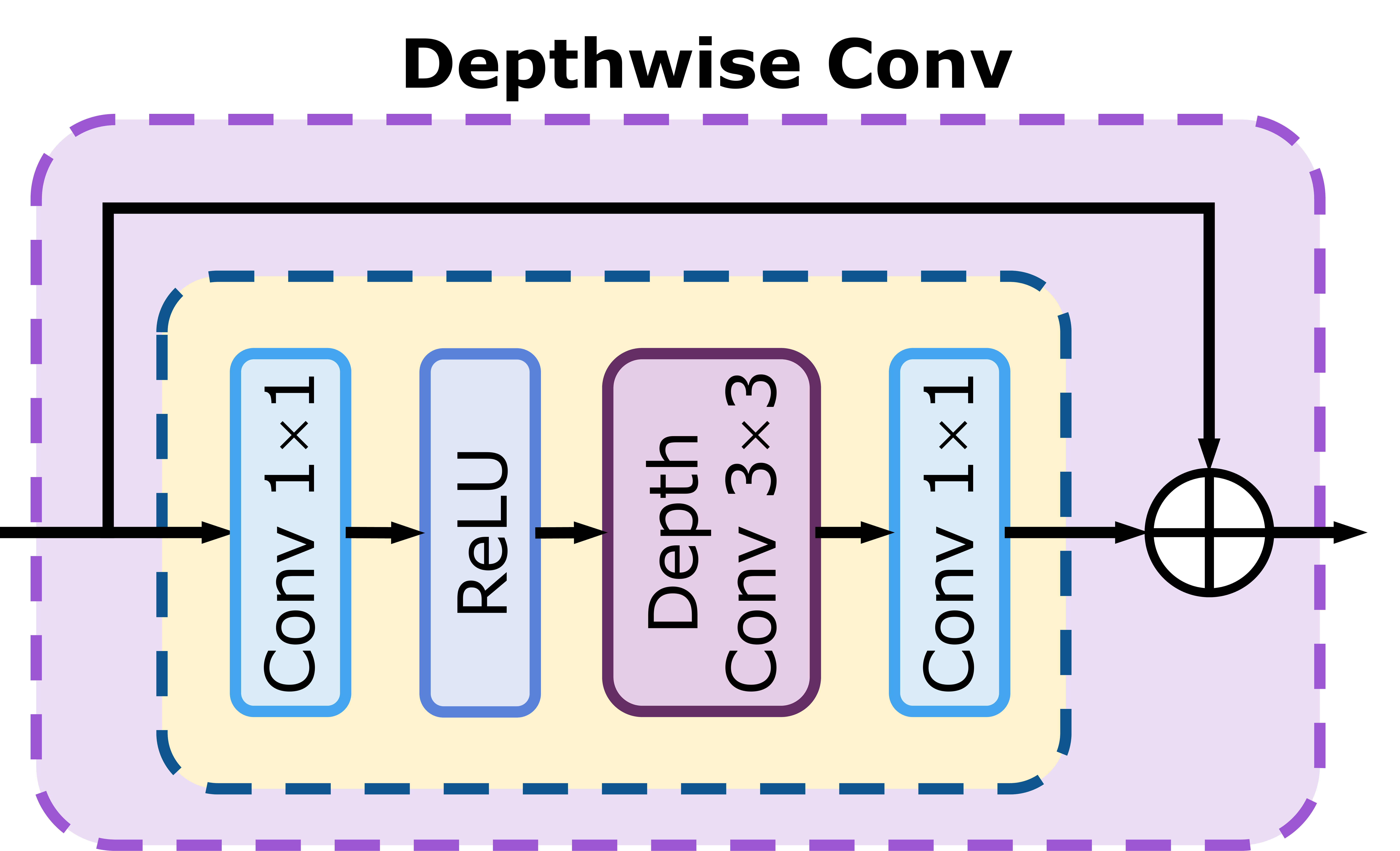}
      \caption{Depthwise Convolution Block.}
      \label{fig:short-b}
    \end{subfigure}
    \vfill
    \begin{subfigure}{0.9999\linewidth}
      \includegraphics[width=0.9999\linewidth]{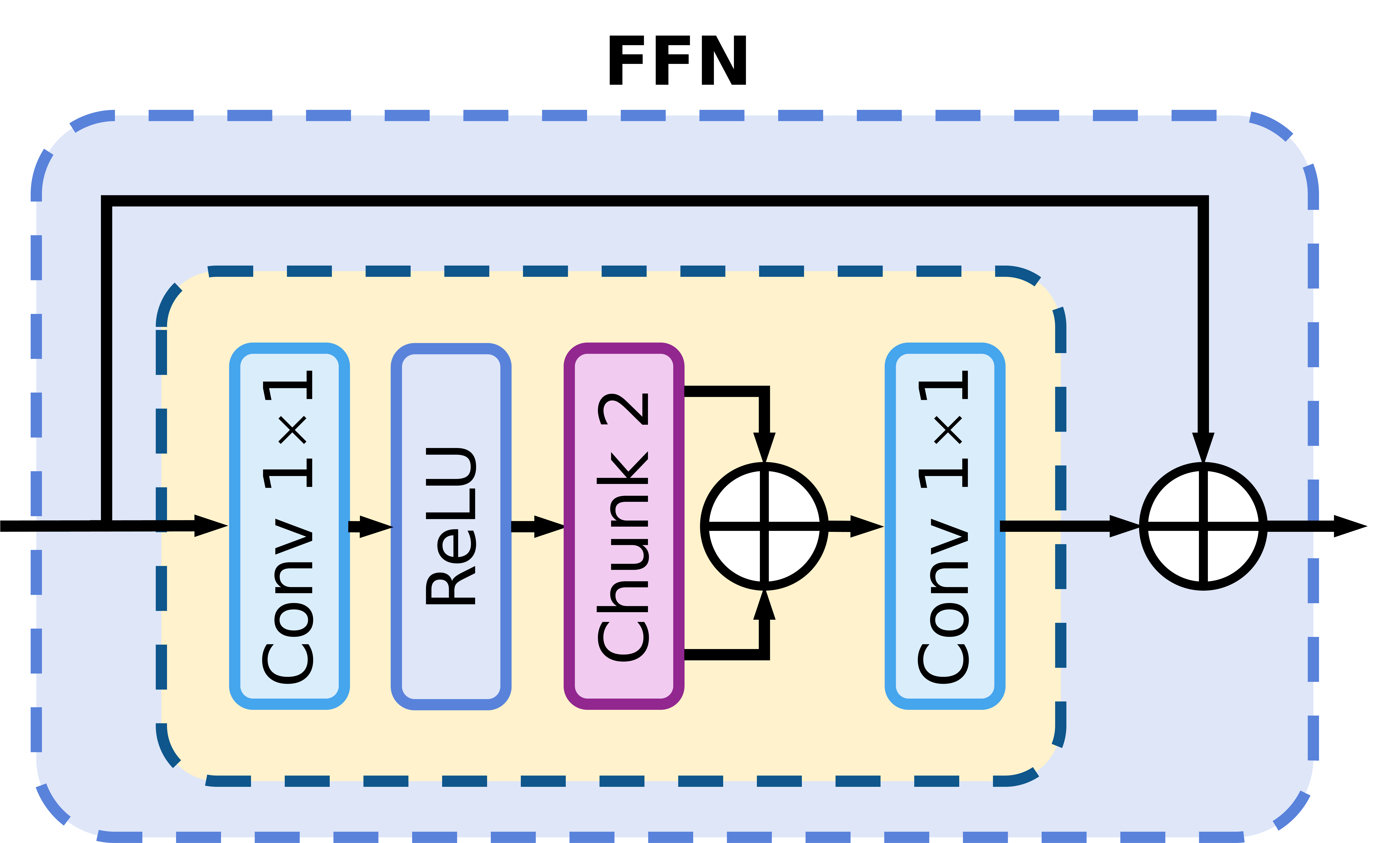}
      \caption{Feed-forward Network.}
      \label{fig:short-c}
    \end{subfigure}
  \end{subfigure}
  \caption{(a) Overview of the proposed ProGIC. Each down-/up-sampling stage consists of a stack of $M$ depthwise convolution blocks and a feed-forward network (FFN). The blocks in $g_s(\cdot)$ are modified with feature modulation, as described in \cref{sec:3.2}. (b) Depthwise convolution block. “Depth conv” denotes a depthwise convolution, while others are pointwise convolutions. (c) FFN architecture, where “Chunk-2” splits the tensor into two equal parts along the channel dimension.}
  \label{fig:short2}
\end{figure*}
\subsection{Learned Image Compression}

\paragraph{Non-Generative Codecs.} Although traditional codecs such as JPEG~\cite{jpeg} and VVC~\cite{vtm} have been highly successful, recent learned image compression (LIC) methods have shown great potential. Pioneering work~\cite{lic-balle2018variational} introduced variational autoencoders (VAEs)~\cite{vae} for image compression. Several studies~\cite{lic-attention, lic-elic, lalic} improved the transform coding~\cite{trans}, while others focused on entropy modeling~\cite{lic-lu2025learned, hpcm}.

Early works in image compression primarily supported one bitrate in one model, whereas subsequent studies have explored more flexible architectures, such as progressive decoding~\cite{lic-progdtd, lic-dpict, lic-ctc, diffusion-prog}. DCVC-RT~\cite{dcvc-rt} also explored practical architectures for real-time coding.

\paragraph{Generative Codecs.} Nevertheless, the above approaches are optimized at the pixel level, leading to a gap in perceptual quality~\cite{perception}. Several works~\cite{gic-1, lic-poelic} attempted to bridge this gap. HiFiC~\cite{hific} leveraged GANs~\cite{gan} to generate high-fidelity reconstructions. MS-ILLM~\cite{ms-illm} further refined the discriminator architecture to enhance distributional fidelity.

Several studies investigate using VQ-GAN~\cite{vq-gan} for learned image compression. Mao~et~al.~\cite{vq-mao2024extreme} used a fine-tuned VQ-GAN for image compression at low bitrates. GLC~\cite{vq-gic} applies additional compression to the latents produced by VQ. Control-GIC~\cite{cgic} designed codebooks at different granularities to achieve multirate compression. DLF~\cite{vq-DLF} utilized VAE for detail compression and VQ-GAN for semantic compression, achieving high perceptual quality at extremely-low bitrates.

Recent works focus on diffusion-based~\cite{ddpm} generative methods for image compression. CDC~\cite{diffusion-cdc} used a conditional diffusion model as the decoder to reconstruct images. PerCo~\cite{diffusion-perco} adopted a discrete codebook for better representation. DiffEIC~\cite{diffeic} leveraged pretrained Stable Diffusion~\cite{stable-diffusion} to reconstruct images. The recent work OSCAR~\cite{oscar} introduced one-step diffusion, significantly accelerating runtime. More works~\cite{diffusion-StableCodec, diffusion-xue2025one} also utilized one-step diffusion for image compression at extremely-low bitrates. Additionally, recent studies~\cite{llm-gao2025exploring, llm-li2024misc} explored using large language models for image compression.

Despite recent progress, research on flexible GICs remains limited: while several methods~\cite{vq-gic, cgic, oscar} offer multirate compression, none address progressive decoding.

\subsection{Vector Quantization in Latent Space}

Vector quantization (VQ)~\cite{vq-1} was widely used in image compression several decades ago. However, due to the high complexity of VQ and the weak decorrelation capabilities of linear transforms, VQ received less attention than non-uniform quantization in traditional codecs~\cite{jpeg}. In neural image compression, uniform quantization techniques have also gained significant attention~\cite{lic-balle2018variational,lic-mlic,hpcm,lalic}.

Nevertheless, VQ in VAEs can address the issue of ``posterior collapse''~\cite{vq-vae}, making it a focal point in generative models. NVTC~\cite{vq-feng2023nvtc} used VQ in every layer of the encoder and decoder. Hao~et~al.~\cite{vq-lattice} introduced lattice VQ for multirate compression. Many generative codecs~\cite{vq-mao2024extreme, vq-gic, cgic, vq-DLF} also utilized VQ for latent representations.

Since the representational ability of a single codebook is limited, Zhu et al.~\cite{vq-zhu2022unified} downsampled latent features layer by layer, applying VQ with multiple codebooks at each layer and using the vectors from the next layer as priors for encoding the previous layer. As for discrete speech representations, SoundStream~\cite{speech-soundstream} introduced a residual vector quantization (RVQ) architecture that employs multiple codebooks to successively encode the residuals of the original latent and prior stages, enabling finer-grained latent representations. DAC~\cite{speech-dac} further improved this RVQ architecture by introducing latent feature projection.
\section{Methods}

\subsection{Overall Architecture with RVQ}
\label{sec:3.1}
\cref{fig:short2} illustrates the overall architecture of ProGIC. The input RGB image $\bm{x} \in  \mathbb{R}^{3\times H \times W}$ is first downsampled by a factor of 8 via pixel unshuffling and then encoded into latent features $\bm{y}$ with another 2$\times$ downsampling by the analysis transform $g_a(\cdot)$, which is
\begin{equation}
  \bm{y} = g_a(\bm{x}).
  \label{eq:analysis}
\end{equation}

The latent $\bm{y}$ is then quantized by RVQ to obtain $\hat{\bm{y}}$. The synthesis transform $g_s(\cdot)$ reconstructs the image $\hat{\bm{x}}$ from $\hat{\bm{y}}$:
\begin{equation}
  \hat{\bm{x}} = g_s(\hat{\bm{y}}).
  \label{eq:synthesis}
\end{equation}

For RVQ, let $Q(\cdot, \bm{C}_i)$ denotes the quantization operation of the $i$-th codebook, with a total of $N$ codebooks. The first codebook $\bm{C}_1$ quantizes the originafl latent features $\bm{y}$ to obtain the base quantized latent $\hat{\bm{y}}_1$. The residual $\bm{r}_1$ is then computed as the difference between $\bm{y}$ and $\hat{\bm{y}}_1$:
\begin{equation}
  \hat{\bm{y}}_1 = Q(\bm{y}, \bm{C}_1), \quad
  \bm{r}_1 = \bm{y} - \hat{\bm{y}}_1.
  \label{eq:rvq1}
\end{equation}

Subsequently, $\bm{r}_1$ is quantized and a new residual is computed. This process continues up to the $N$-th codebook $\bm{C}_N$, each stage quantizing the current residual. For $1 \le i < N$,
\begin{equation}
  \hat{\bm{r}}_i = Q(\bm{r}_{i}, \bm{C}_{i+1}), \quad
  \bm{r}_{i} = \bm{y} - \hat{\bm{y}}_1 - \sum_{j=1}^{i-1}\hat{\bm{r}}_{j}.
  \label{eq:rvq2}
\end{equation}

The process to get quantized latent features $\hat{\bm{y}}$ is:
\begin{equation}
  \hat{\bm{y}} = \hat{\bm{y}}_1 + \sum_{i=1}^{N-1} \hat{\bm{r}}_i.
  \label{eq:rvq3}
\end{equation}

The indices of the selected codewords from each codebook are transmitted to the decoder, which retrieves the corresponding codewords from the shared codebooks. For a model using $N$ codebooks, each containing $2^L$ vectors, the bit per pixel (BPP) is given by $\text{BPP} = \frac{N\times L}{16 \times 16}$,
where 16 is the spatial downsampling ratio of original images. 

We do not use entropy coding on these indices because we observe only a 0.9\% bitrate reduction when applying range coding~\cite{range-coding}, as discussed in \cref{sec:ab}.



Compared with prior multi-codebook designs~\cite{vq-vae-2, vq-zhu2022unified} which are typically instantiated with multi-scale architectures to obtain latents at different levels, RVQ formulation operates on a single-resolution latent and refines it through stage-wise residual vector additions. This results in an efficient implementation. Moreover, we show that RVQ is inherently well-suited for progressive decoding, whereas prior multi-codebook methods do not support. A detailed analysis of the codebook properties is provided in the supplementary material.

\subsection{Lightweight Backbone with Attention and Feature Modulation}
\label{sec:3.2}
Since the RVQ structure in ProGIC only requires limited inexpensive vector addition, the overall complexity is mainly determined by $g_a(\cdot)$ and $g_s(\cdot)$. We follow the lightweight backbone in DCVC-RT~\cite{dcvc-rt} and further enhance it with attention modules and feature-modulation layers to improve representational capacity. As illustrated in \cref{fig:short2}(a), we adopt $g_a(\cdot)$ and $g_s(\cdot)$ at a single resolution of $\frac{H}{8}\times\frac{W}{8}$. We stack depthwise convolution blocks and feed-forward networks (FFNs) $M$ times as the main components of $g_a(\cdot)$ and $g_s(\cdot)$, which significantly reduces complexity compared with ResBlocks~\cite{lic-elic}. \cref{fig:short2}(b) and \cref{fig:short2}(c) show the details of the depthwise convolution block and the FFN, respectively. We replace the WSiLU activation~\cite{dcvc-rt} in the original implementation with ReLU~\cite{relu} for faster computation on CPUs.

Furthermore, we observe that small $M$ leads to insufficient spatial aggregation in the depthwise convolution blocks, whereas increasing $M$ significantly raises computational complexity. To address this trade-off, we introduce an attention module after the downsampling layer in $g_a(\cdot)$ and before the upsampling layer in $g_s(\cdot)$. The module architecture follows \cite{lic-attention}, but we replace the original ResBlocks by stacking $k=3$ depthwise convolution blocks and FFNs to reduce complexity. This design compensates for weak spatial aggregation at small $M$ while capturing long-range dependencies, thereby improving representational capacity at low computational cost.

\begin{figure}[t]
  \centering
  \includegraphics[width=0.70\linewidth]{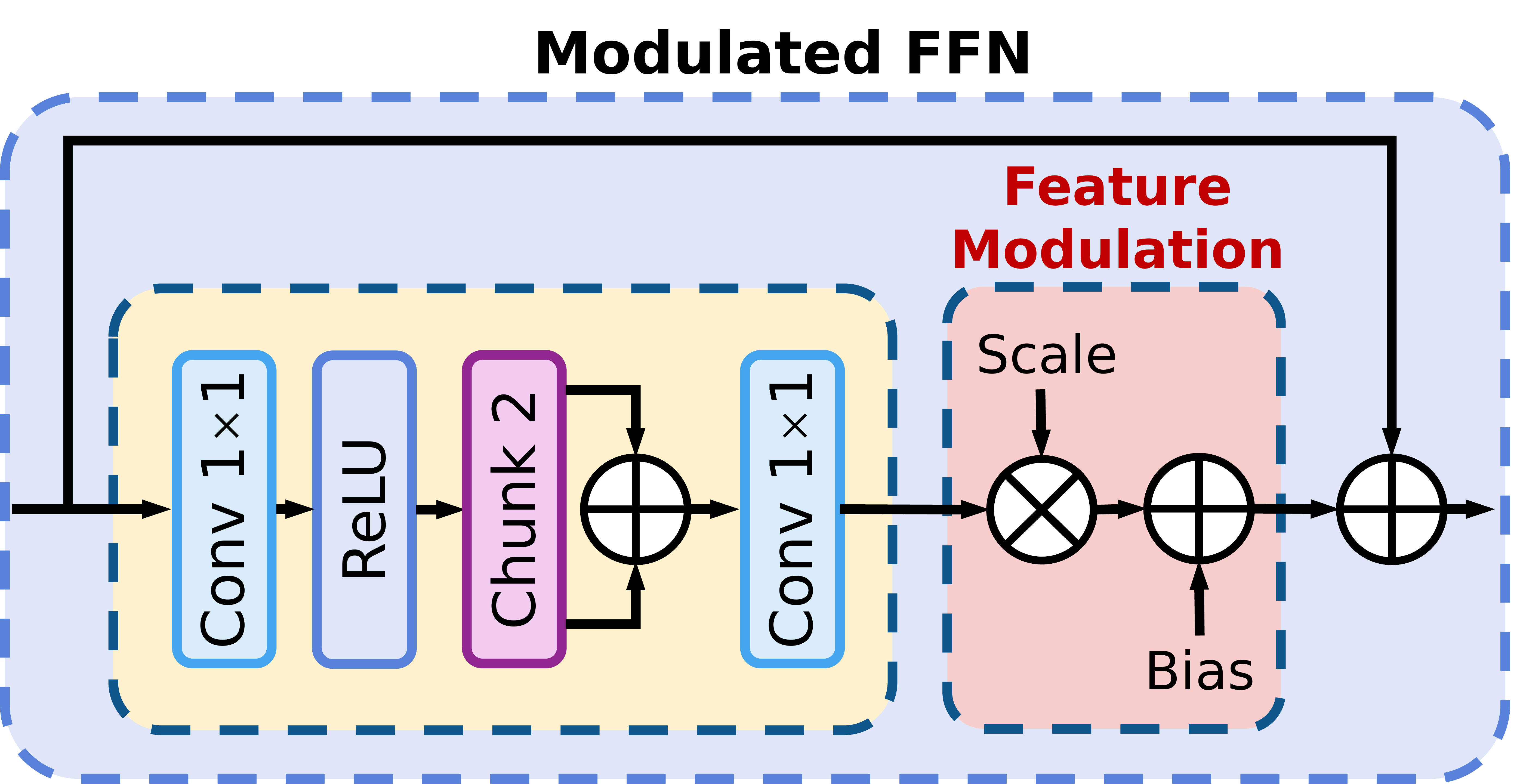}
  \caption{Feature modulation in an FFN: at each progressive decoding stage, stage-specific scale and bias are applied to the features before the residual addition.}
  \label{fig:fm}
\end{figure}

To make the model aware of progressive decoding, we incorporate feature modulation into $g_s(\cdot)$. Specifically, before the residual connection of each depthwise convolution block and FFN in $g_s(\cdot)$, we modulate the features by multiplying a scale factor and adding a bias term. Each progressive stage $i$ (i.e., using the first $i$ codebooks) uses distinct scale and bias parameters. This design introduces minimal computational overhead while effectively improving performance under progressive decoding. \cref{fig:fm} shows the feature modulation mechanism through an FFN example.

\subsection{Learning Strategy for Progressive Decoding}
\label{sec:3.3}
As illustrated in \cref{fig:prog_illustrate}, RVQ is inherently suited for progressive decoding. The core idea is simple: using only the first $i$ codebooks yields the $i$-th-stage reconstruction. During training, we iterate over $i \in {1,\dots,N}$ and, for each $i$, compute the reconstruction using only the first $i$ codebooks. We accumulate the corresponding losses across all $i$ and perform a single backpropagation step. To enhance the final reconstruction quality and balance the contribution of different progressive stages, we weight the loss at each stage $i$ by a coefficient $\lambda_i$.

Following common practice in GICs, our supervised training objective combines reconstruction loss, perceptual loss~\cite{lpips}, adversarial loss~\cite{gan}, and codebook loss~\cite{vq-vae}. The total training loss can be formulated as 

\begin{equation}
  \begin{aligned}
    \mathcal{L} = \sum_{i=1}^{N} \lambda_i \Big( 
      & \lVert \bm{x} - \hat{\bm{x}}_i\rVert_1 
      + \lambda_{\text{per}} \, \mathcal{L}_{\text{per}}(\bm{x}, \hat{\bm{x}}_i) \\
      & + \lambda_{\text{adv}} \, \mathcal{L}_{\text{adv}}(\bm{x}, \hat{\bm{x}}_i) 
      + \lambda_{\text{cb}} \, \mathcal{L}_{\text{cb}}^{i}
    \Big).
  \end{aligned}
\end{equation}

Here, $\hat{\bm{x}}_i$ denotes the reconstruction at the $i$-th stage, obtained using the first $i$ codebooks. $\mathcal{L}_{\text{per}}$ is the LPIPS loss computed using the VGG network~\cite{vgg}. $\mathcal{L}_{\text{adv}}$ is the adaptive PatchGAN adversarial loss~\cite{vq-gan}. The codebook loss $\mathcal{L}_{\text{cb}}^{i}$ consists of commitment loss and codebook update loss~\cite{vq-vae}. $\lambda_i$ can be calculated as
\begin{equation}
  \lambda_i = \left\{
    \begin{aligned}
      \frac{p}{N-1}, & \quad i < N \\
      1 - p, & \quad i = N
    \end{aligned}
  \right.,
\end{equation}
where $p$ is a weighting ratio that controls the trade-off between the final reconstruction and intermediate stages, ensuring $\sum_{i=1}^{N}\lambda_i=1$ for any $p\in[0,1]$. A smaller $p$ places more weight on the final reconstruction. When $p=0$, all weight is assigned to the final stage ($i=N$). In our experiments, we set $p = 0.5$ and analyze its effect in \cref{sec:ablation_prog}.


\begin{figure*}[t]
  \centering
  \includegraphics[width=0.99\linewidth]{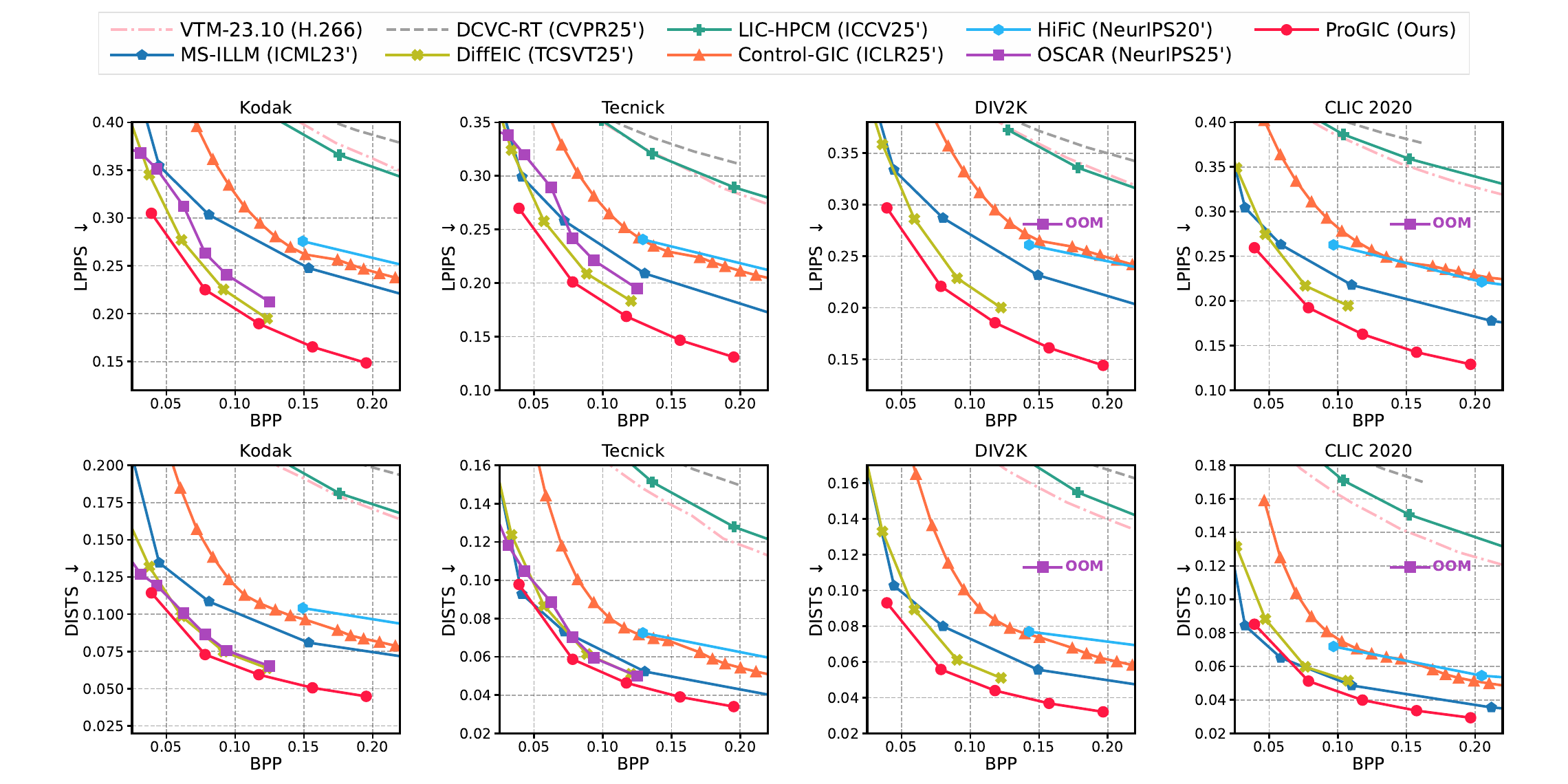}
  \caption{Rate-distortion performance on the Kodak, Tecnick, DIV2K, and CLIC2020-Professional datasets, evaluated with LPIPS and DISTS vs. BPP. Curves closer to the lower-left  are better, indicating better quality at the same compression ratio.  ``OOM'' denotes out-of-memory under the official evaluation environment.}
  \label{fig:RD-curve}
\end{figure*}

\begin{table*}[t]
  \caption{BD-rate and computational complexity on the Kodak dataset, evaluated on an NVIDIA A100 GPU. Best results are in \textbf{bold}. Second best are \underline{underlined}. A dash (--) denotes unavailable results. “Enc./Dec.” indicates encoding/decoding time per image.}
  \label{tab:bd_table}
  \centering
  \begingroup
  \setlength{\tabcolsep}{5.5pt} 
  \begin{tabular}{@{}lrrrrrrrr@{}}
    \toprule
    \multirow{2}{*}{Method} &
    \multirow{2}{*}{Enc.(ms)} &
    \multirow{2}{*}{Dec.(ms)} &
    \multirow{2}{*}{FLOPs(G)} &
    \multirow{2}{*}{Params(M)} &
    \multicolumn{4}{c}{BD-rate (LPIPS)} \\
    \cmidrule(lr){6-9}
    & & & & & Kodak & Tecnick & DIV2K & CLIC 2020 \\
    \midrule
    VTM-23.10 (Intra)~\cite{vtm}& $>$9999       & 150.30       & --       & --  & 313.84\%  & 295.19\%  & 285.10\%  & 498.64\%  \\
    LIC-HPCM~\cite{hpcm}        & 62.37   & 82.88   & 732.47  & 68.50   & 274.50\%  & 305.42\%  & 267.18\%  & 745.04\%  \\
    DCVC-RT (Intra)~\cite{dcvc-rt} & 14.09   & 17.08   & 382.98 & 45.65    & 393.72\%  & 329.44\%  & 349.39\%  & 584.41\%  \\
    \midrule
    HiFiC~\cite{hific}           & 526.51  & 1408.60 & 599.51  & 181.60  & 45.82\% & 68.66\% & 46.36\% & 86.45\% \\
    Control-GIC~\cite{cgic}     & 103.56  & 436.26  & 5816.37 & 130.36  & 33.36\% & 68.83\% & 73.77\% & 136.25\% \\
    MS-ILLM~\cite{ms-illm}         & 165.38  & 147.79  & 599.52  & 181.40  & 0.00\%    & 0.00\%    & 0.00\%    & 0.00\%    \\
    DiffEIC~\cite{diffeic}        & 210.18 & 4661.74 & 57339.93  & 1379.50      & -37.71\% & -9.96\% & -15.76\% & 4.34\% \\
    OSCAR~\cite{oscar}           & 53.04   & 167.56  & 6485.61 & 1009.30 & -37.31\% & -5.76\% & -- & -- \\
    \midrule
    ProGIC-s (Ours) & \textbf{6.13}    & \textbf{7.66}    & \textbf{108.28}  & \textbf{14.44}   & \underline{-52.73\%} & \underline{-37.86\%} & \underline{-46.02\%} & \underline{-42.76\%} \\
    ProGIC (Ours)   & \underline{7.64}    & \underline{10.99}   & \underline{333.38}  & \underline{33.11}   & \textbf{-58.83\%} & \textbf{-45.53\%} & \textbf{-51.77\%} & \textbf{-51.13\%} \\
    \bottomrule
  \end{tabular}
  \endgroup
\end{table*}

\section{Experiments}
\label{sec:4}
\subsection{Experimental Setup}

\paragraph{Implementation Details.}
To build the model, we set $C_1$ to 368 in the base model and 256 in the small model. The latent feature dimension $C_2$ is fixed at 256 for both variants. 
The encoder comprises $M = 8$ Depthwise Conv blocks in the base model and $M = 4$ blocks in the small model, while the decoder uses 14 and 8 layers, respectively. 
We adopt an expansion ratio $r = 4$ in FFNs. 
We employ $N = 5$ codebooks, each containing $2^{L} = 1024$ embedding vectors, enabling progressive decoding at BPPs of $\{0.0391, 0.0781, 0.1172, 0.1562, 0.1953\}$ in a single model.
We give more models with different BPP ranges in the supplementary material.

\paragraph{Training Details.}
We train our model on the full ImageNet dataset~\cite{imagenet}. 
In each epoch, we sample 1\% of the images and extract $256\times256$ random crops, then apply data augmentation with random horizontal flipping. Optimization is performed using the Adam optimizer~\cite{adam} with $\beta_1 = 0.5$, $\beta_2 = 0.9$, and an initial learning rate of $10^{-4}$, which is decayed to $10^{-5}$ after 1.5M iterations. The model is trained for 2M iterations with a batch size of 16 on one NVIDIA A100 GPU with peak memory usage of 12.4GB.

\paragraph{Evaluation Datasets and Metrics.}
We evaluate the proposed model on four datasets: (1) Kodak~\cite{Kodak}, containing 24 images with $768 \times 512$ resolution, (2) Tecnick~\cite{tecnick}, containing 100 images at $1200\times1200$ resolution, (3) DIV2K~\cite{div2k}, containing 100 images at typically $2048\times1024$ resolution, and (4) CLIC 2020 Professional~\cite{clic2020}, consisting of 250 images with resolutions up to 2K resolution. Following prior work~\cite{hific, ms-illm, diffeic}, we use LPIPS~\cite{lpips} and DISTS~\cite{dists} as primary metrics. Results for PSNR, MS-SSIM~\cite{ms-ssim}, and CLIP-IQA~\cite{clip-iqa} are provided in the supplementary material.

\subsection{Rate-Distortion Performance}

\paragraph{Comparison with Previous Methods.}
We compare the rate-distortion (R--D) performance of our method with previous approaches. For non-generative codecs, we include the traditional VTM-23.10~\cite{vtm} and the learned codecs LIC-HPCM~\cite{hpcm} and DCVC-RT~\cite{dcvc-rt}. For GICs, we compare with the GAN-based HiFiC~\cite{hific} and MS-ILLM~\cite{ms-illm}, the VQ-based Control-GIC~\cite{cgic}, and the diffusion-based DiffEIC~\cite{diffeic} and OSCAR~\cite{oscar}. For fair comparison, we use the official pretrained models provided by the authors, and run FP32 models with a batch size of 1 on the same server with a single NVIDIA A100 GPU and an AMD EPYC 7763 CPU for all methods. The main results are shown in \cref{fig:RD-curve}. We observe that the official settings for OSCAR lead to out-of-memory errors on the DIV2K and CLIC 2020 datasets, as it requires more than 80 GB of memory for processing a single image.

As summarized in \cref{tab:bd_table}, ProGIC achieves the best performance across the evaluated datasets and metrics. Notably, ProGIC achieves significant BD-rate reductions of 58.85\%, 45.35\%, 51.77\%, and 51.13\% on the Kodak, Tecnick, DIV2K, and CLIC 2020 datasets on LPIPS, respectively. BD-rate for DISTS is provided in the supplementary material, where ProGIC also achieves previous performance across all datasets.

\begin{figure*}[t]
  \centering
  \includegraphics[width=0.90\linewidth]{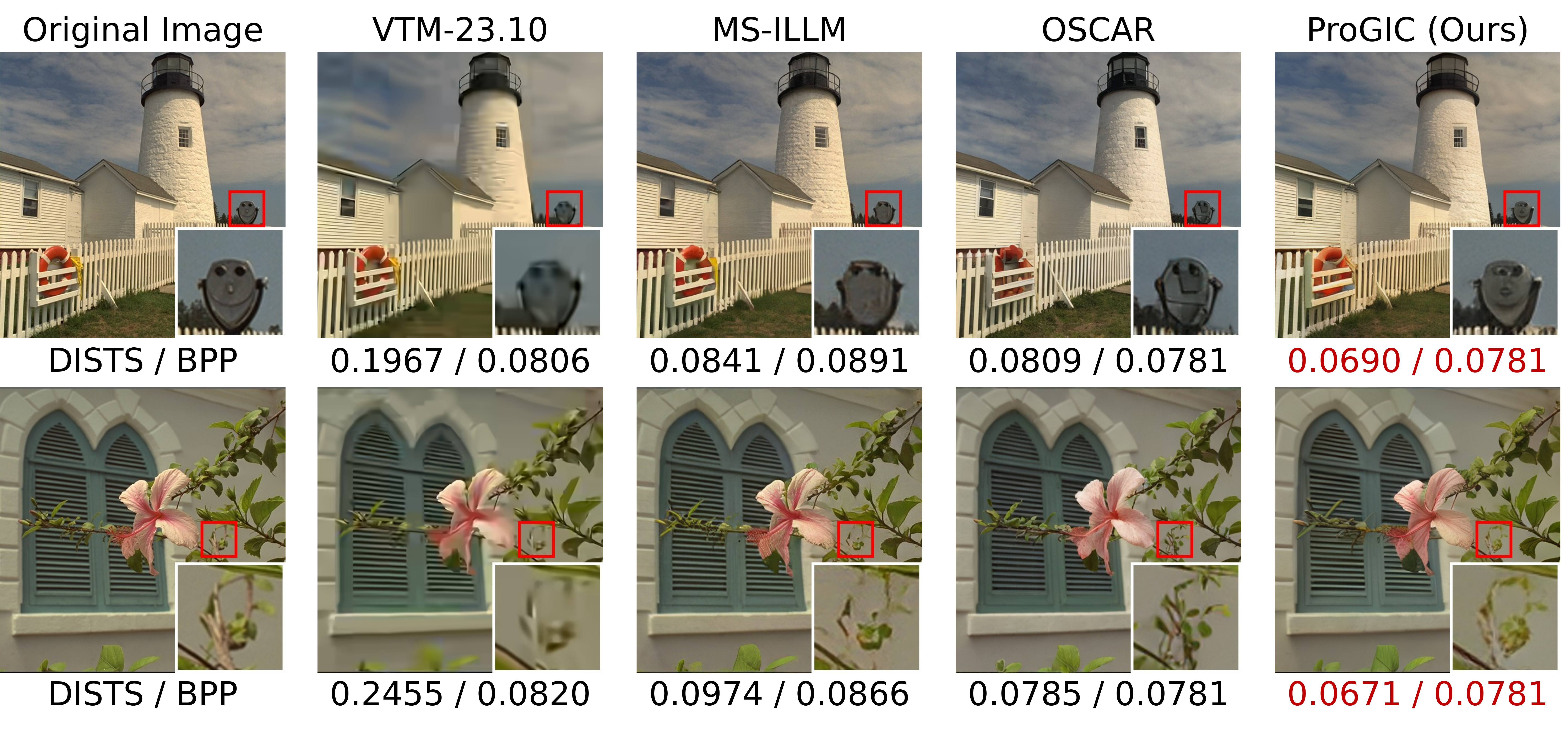}
  \caption{Visualization of reconstructed images from different methods on Kodak. Values denote DISTS / bpp. Lower DISTS indicates better perceptual quality. Lower bpp indicates higher compression.}
  \label{fig:vis}
\end{figure*}

\cref{fig:vis} shows visual comparisons of reconstructed images from different methods at similar bitrates on the Kodak dataset. In the first row, the smiling face is noticeably blurred in VTM-23.10 and MS-ILLM. OSCAR hallucinates facial features that do not exist in the original image, while ProGIC faithfully recovers the original smile. In the second row, OSCAR produces a result different from the original branch structure, whereas ProGIC preserves both the color and structural details of the branches. These observations explain superior DISTS scores of ProGIC, as DISTS is sensitive to structural and textural differences~\cite{dcvc-rt}.

\begin{table}[t]
  \caption{Laptop AMD Ryzen 7840HS CPU runtimes (ms) for image encoding and decoding across different resolutions. Best results are in \textbf{bold}. Second-best are \underline{underlined}.}
  \label{tab:cpu_time}
  \centering
  \begin{tabular}{@{}lrrrr@{}}
    \toprule
    \multirow{2}{*}{Method} & \multicolumn{2}{c}{256$\times$256} & \multicolumn{2}{c}{512$\times$512} \\
    \cmidrule(lr){2-3} \cmidrule(lr){4-5}
     & Enc. & Dec. & Enc. & Dec. \\
    \midrule
    VTM-23.10 (Intra)~\cite{vtm} & $>$10s & 330 & $>$10s & \underline{456} \\
    DCVC-RT (Intra)~\cite{dcvc-rt} & 247 & 187 & 981 & 725 \\
    MS-ILLM~\cite{ms-illm} & 121 & 368 & 507 & 1352 \\
    DiffEIC~\cite{diffeic} & -- & $>$10s & -- & $>$10s \\
    OSCAR~\cite{oscar} & 805 & 2530 & 3429 & 9519 \\
    \midrule
    ProGIC (Ours) & \underline{76} & \underline{124} & \underline{297} & 515 \\
    ProGIC-s (Ours) & \textbf{34} & \textbf{50} & \textbf{107} & \textbf{184} \\
    \bottomrule
  \end{tabular}
\end{table}

\subsection{Complexity Analysis}
\paragraph{Complexity on GPUs.}
, we report encoding and decoding runtimes (in milliseconds), the number of floating-point operations (FLOPs, in gigaflops), and the number of model parameters (in millions) on the same server mentioned above. More detailed runtime results for different image resolutions are provided in the supplementary material.

As summarized in \cref{tab:bd_table}, ProGIC achieves significantly lower complexity than previous methods. Our base model is over $5\times$ faster in encoding and $10\times$ faster in decoding than OSCAR on one GPU. It is less time-consuming than the fastest non-generative method DCVC-RT as well. Our small model (ProGIC-s) achieves the best runtime, while still outperforming all other methods in R--D performance. This highlights a deployment-oriented insight: competitive GIC can be achieved without expensive tokenizers, which is a major practical bottleneck.

\paragraph{Complexity on Laptop CPUs.}
Moreover, we measure FP32 CPU runtimes on the same laptop equipped with an AMD Ryzen 7840HS. As shown in \cref{tab:cpu_time}, ProGIC-s is significantly faster than other methods on CPU. It also outperforms the traditional codec VTM-23.10~\cite{vtm} on laptop CPU, highlighting its practical utility.

\paragraph{Complexity on Mobile Phone CPUs.}
We provide the runtime of ProGIC-s on Android CPU devices. As shown in~\cref{tab:android_time}, we evaluate ProGIC-s on mobile phones equipped with the Snapdragon 870 (released in 2021) and MediaTek Dimensity 8000 (released in 2022). Both chips are substantially outperformed by mainstream mobile processors in today. Results show that ProGIC-s achieves feasible encoding and decoding on these limited mobile devices, demonstrating its practicality.
We provide a use case with ProGIC in the supplementary material as well.  

\begin{table}[t]
  \caption{Mobile phones CPU runtimes (s) for image encoding and decoding with ProGIC-s across different resolutions.}
  \label{tab:android_time}
  \centering
  \begin{tabular}{@{}lrrrr@{}}
    \toprule
    \multirow{2}{*}{CPU} & \multicolumn{2}{c}{256$\times$256} & \multicolumn{2}{c}{512$\times$512} \\
    \cmidrule(lr){2-3} \cmidrule(lr){4-5}
     & Enc. & Dec. & Enc. & Dec. \\
    \midrule
    Snapdragon 870 & 0.563 & 0.675 & 2.498 & 3.827 \\
    MTK Dimensity 8000 & 0.559 & 0.715 & 2.346 & 3.658 \\
    \bottomrule
  \end{tabular}
\end{table}

\subsection{Ablation Studies}
\label{sec:ab}
In this section, we present ablation studies to validate the contribution of each proposed component. For a clearer comparison, we evaluate BD-rate on the Kodak dataset using DISTS as the quality metric. We run all ablation experiments with 1M iterations for efficiency.

\paragraph{Proposed progressive quantization.}
Existing progressive codecs~\cite{lic-dpict,lic-ctc,lic-progdtd,diffusion-prog} are not optimized for perceptual quality and are thus incomparable to ProGIC. To ensure a fair and comprehensive comparison, we retrain the non-generative progressive codec ProgDTD~\cite{lic-progdtd} using the same dataset and training protocol as ProGIC. ProgDTD achieves progressive decoding without introducing additional components.We use the training hyperparameter $\lambda \in \{2,4,8\}$. We find $\lambda=2$ yields a bitrate range comparable to ProGIC. 

\cref{tab:1} presents a full ablation study of the proposed progressive quantization. We fix the lightweight backbone as ``Base'', while the other variants add them individually or together.
As shown in \cref{tab:1}, although ``Base + ProgDTD (PSNR)'' has poor BD-rate, ``Base + ProgDTD (LPIPS)'' optimized with the same objective as ProGIC already outperforms MS-ILLM with a 10.28\% gain, confirming the strength of our lightweight backbone for GIC.
Moreover, ``Base + RVQ'' significantly improves BD-rate by 37.82\% and delivers a $5\times$ speedup over ``Base + ProgDTD (LPIPS)'', confirming the effectiveness of RVQ that brings strong perceptual quality and low coding latency.

\paragraph{Proposed Modules and Entropy Coding.}
As shown in \cref{tab:1}, Feature modulation incurs a negligible change in FLOPs but yields a 2.42\% BD-rate reduction. The attention module contributes a more substantial improvement of 11.41\% BD-rate reduction. Combining both modules achieves a 14.90\% BD-rate reduction with a minimal increase in encoding and decoding latency. 

Additionally, we conduct experiments to evaluate the effect of entropy coding. The range coding~\cite{range-coding} is applied, consistent with previous learned image codecs~\cite{lic-balle2018variational, lic-elic, lalic, hpcm}. We train the prior distribution on the whole ImageNet dataset and evaluate the BD-rate reduction on the Kodak dataset.
We find that applying entropy coding to the quantized indices only provides a marginal 0.90\% BD-rate reduction. This is because the distribution of indices show little statistical redundancy. Full R--D curve is provided in the supplementary material.


\begin{table}[t]
  \caption{Ablation study on the effect of different components and entropy coding. ``Base'' is the lightweight backbone, and ``ProgDTD'' is the baseline progressive quantization method.}
  \label{tab:1}
  \centering
  \begingroup
  \setlength{\tabcolsep}{2.9pt} 
  \begin{tabular}{@{}lrrr@{}}
    \toprule
     & BD-rate & Enc.(ms) & Dec.(ms)\\
    \midrule
    MS-ILLM                         & 0.0 \% & 165.38 & 147.79 \\
    \midrule
    Base + ProgDTD (PSNR)                       & 487.18 \% & 39.31 & 52.46 \\
    Base + ProgDTD (LPIPS)                      & -10.28\% & 38.37 & 51.38 \\
    \midrule
    Base + RVQ (LPIPS)                          & -48.10\%   & 6.23 & 9.33 \\
     + Attention                                & \underline{-56.00\%}   & 7.70 & 10.61 \\
     + Attention + Modulation                   & \textbf{-57.57\%}   & 7.70 & 10.62 \\
     \midrule
    + Entropy Coding          & -49.0\% & -- \\
    \bottomrule
  \end{tabular}
  \endgroup
\end{table}

\begin{figure}[t]
  \centering
  \includegraphics[width=0.90\linewidth]{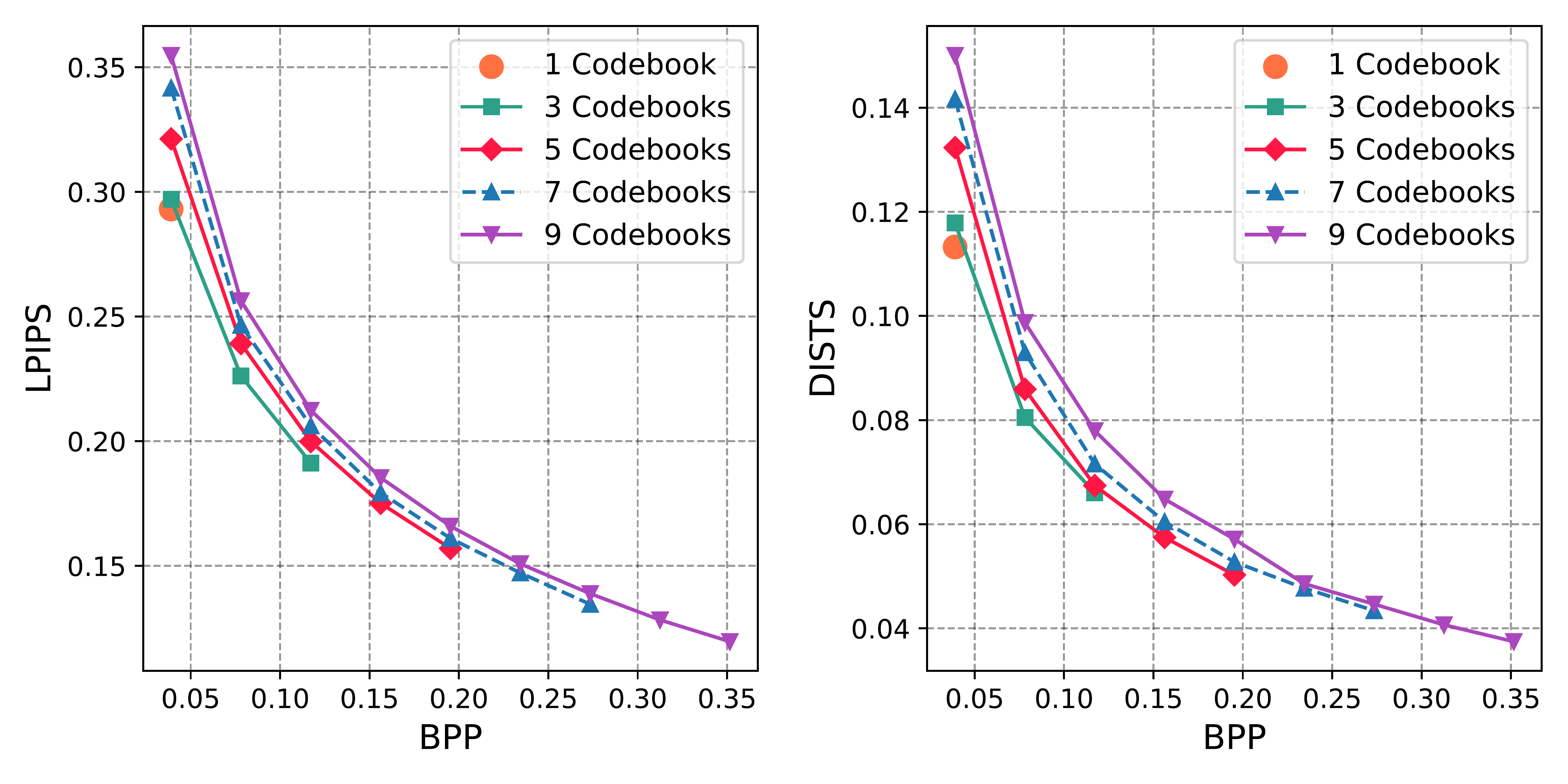}
  \caption{R--D performance with different codebook numbers.}
  \label{fig:codebook}
\end{figure}

\paragraph{Codebook Numbers.}
\label{sec:codebook_num}
As shown in \cref{fig:codebook}, we analyze how the number of codebooks $N$ affects R–D performance. Increasing $N$ can result to a larger bitrate range. However, the quality of intermediate reconstructions for progressive decoding drops. When targeting a wider bitrate range, this leads to poorer preview quality at lower bitrates. This limitation is also observed in~\cite{lic-progdtd}. We find that $N=5$ offers a good balance between progressive reconstruction quality and bitrate range, and thus set $N=5$ in our main model.

\begin{figure}[t]
  \centering
  \includegraphics[width=0.90\linewidth]{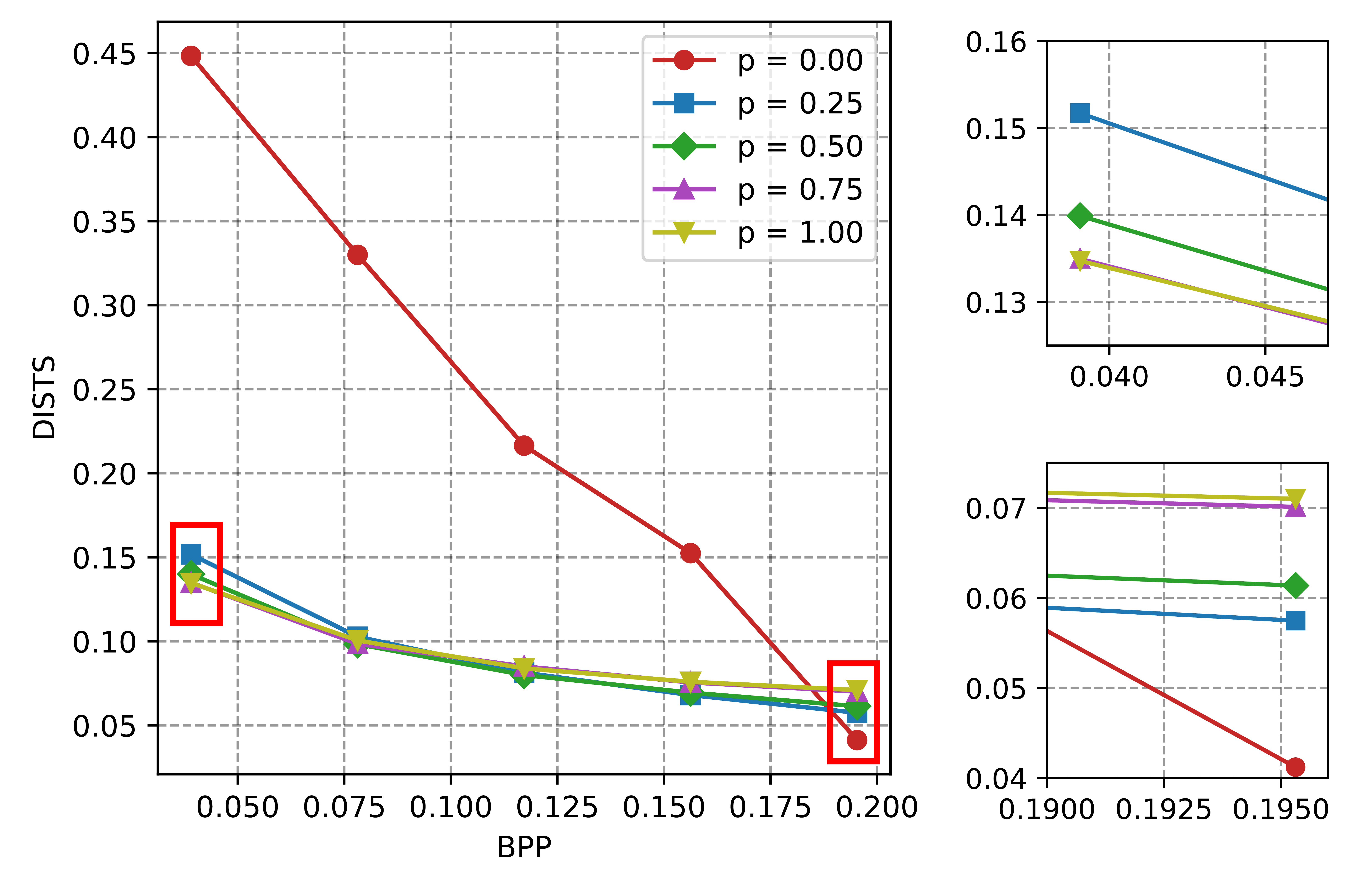}
  \caption{Effect of different weighting ratios $p$ for training. The top-right zoom highlights the low-bitrate region. The bottom-right zoom highlights the high-bitrate region.}
  \label{fig:prog_ab}
\end{figure}

\paragraph{Progressive Decoding.}
\label{sec:ablation_prog}
In \cref{sec:3.3}, we introduce a learning strategy for progressive decoding. The weighting ratio $p$ controls the trade-off between final reconstruction quality and intermediate decoding stages. We conduct experiments with different values of $p$ to analyze its effect, as shown in \cref{fig:prog_ab}. A larger $p$ leads to better performance at low bitrates, while a smaller $p$ improves high-bitrate performance. This is because a larger $p$ encourages the model to focus more on the early stages of reconstruction. We choose $p=0.5$ to balance the performance across all bitrates.

\section{Conclusion}

In this paper, we propose ProGIC, a progressive generative image compression framework built upon RVQ. ProGIC represents the image latent representation as a sum of quantized residual vectors from multiple codebooks. This design enables coarse-to-fine decoding and fast previews from partial bitstreams. We further introduce a lightweight depthwise-separable convolutional backbone augmented with small attention blocks to improve efficiency. As a result, ProGIC is practical on both GPUs and CPU-only mobile devices. Experiments show that ProGIC achieves comparable compression performance on multiple benchmark datasets while being significantly faster.

\section*{Acknowledgment}
This work is supported by the National Key Research and Development Program of China under Grant No. 2023YFB2904300, the National Natural Science Foundation of China under Grant No. 62293484, No. 62441235, and No. 92570204, Beijing Natural Science Foundation (F251001 and L257005), and the Program of Jiangsu Province under Grant No. NTACT-2024-Z-001.

\newpage

{
    \small
    \bibliographystyle{ieeenat_fullname}
    \bibliography{main}

@String(CVPR= {IEEE Conf. Comput. Vis. Pattern Recog.})

@String(ICCV= {Int. Conf. Comput. Vis.})

@String(ICLR = {Int. Conf. Learn. Represent.})

@String(AAAI = {AAAI})

@String(CVPR  = {CVPR})

@String(ICCV  = {ICCV})

@String(ICLR  = {ICLR})

@misc{vtm,
  author       = {{VTM-23.10}},
  year         = {2025},
  howpublished = {\url{https://vcgit.hhi.fraunhofer.de/jvet/VVCSoftware_VTM/}},
  note         = {Accessed: 2025-06-05}
}

@article{jpeg,
  title={The JPEG still picture compression standard},
  author={Wallace, Gregory K},
  journal={Communications of the ACM},
  volume={34},
  number={4},
  pages={30--44},
  year={1991},
  publisher={AcM New York, NY, USA}
}

@misc{kodak,
  author       = {{Kodak Lossless True Color Image Suite}},
  howpublished = {http://r0k. us/graphics/kodak/},
  year={1993}
}

@inproceedings{tecnick,
  title={TESTIMAGES: a Large-scale Archive for Testing Visual Devices and Basic Image Processing Algorithms.},
  author={Asuni, Nicola and Giachetti, Andrea and others},
  booktitle={STAG: Smart Tools and Applications in Computer Graphics},
  pages={63--70},
  year={2014}
}

@inproceedings{div2k,
  title={Ntire 2017 challenge on single image super-resolution: Dataset and study},
  author={Agustsson, Eirikur and Timofte, Radu},
  booktitle={Proceedings of the IEEE conference on computer vision and pattern recognition workshops},
  pages={126--135},
  year={2017}
}

@inproceedings{clic2020,
  title={Workshop and challenge on learned image compression},
  author={CLIC},
  booktitle={Proceedings of the IEEE/CVF Conference on Computer Vision and Pattern Recognition},
  year={2020}
}

@inproceedings{dcvc-rt,
  title={Towards practical real-time neural video compression},
  author={Jia, Zhaoyang and Li, Bin and Li, Jiahao and Xie, Wenxuan and Qi, Linfeng and Li, Houqiang and Lu, Yan},
  booktitle={Proceedings of the IEEE/CVF Conference on Computer Vision and Pattern Recognition},
  pages={12543--12552},
  year={2025}
}

@inproceedings{hpcm,
  title={Learned image compression with hierarchical progressive context modeling},
  author={Li, Yuqi and Zhang, Haotian and Li, Li and Liu, Dong},
  booktitle = {The Twentieth IEEE/CVF International Conference on Computer Vision},
  year      = {2025},
}

@inproceedings{lalic,
  title={Linear Attention Modeling for Learned Image Compression},
  author={Feng, Donghui and Cheng, Zhengxue and Wang, Shen and Wu, Ronghua and Hu, Hongwei and Lu, Guo and Song, Li},
  booktitle={Proceedings of the Computer Vision and Pattern Recognition Conference},
  pages={7623--7632},
  year={2025}
}

@inproceedings{lic-lu2025learned,
  title={Learned Image Compression with Dictionary-based Entropy Model},
  author={Lu, Jingbo and Zhang, Leheng and Zhou, Xingyu and Li, Mu and Li, Wen and Gu, Shuhang},
  booktitle={Proceedings of the Computer Vision and Pattern Recognition Conference},
  pages={12850--12859},
  year={2025}
}

@article{lic-mlic,
  author = {Jiang, Wei and Yang, Jiayu and Zhai, Yongqi and Gao, Feng and Wang, Ronggang},
  title = {MLIC++: Linear Complexity Multi-Reference Entropy Modeling for Learned Image Compression},
  year = {2025},
  issue_date = {May 2025},
  publisher = {Association for Computing Machinery},
  address = {New York, NY, USA},
  volume = {21},
  number = {5},
  issn = {1551-6857},
  doi = {10.1145/3719011},
  journal = {ACM Trans. Multimedia Comput. Commun. Appl.},
  month = may,
  articleno = {142},
  numpages = {25}
}

@inproceedings{lic-elic,
  title={Elic: Efficient learned image compression with unevenly grouped space-channel contextual adaptive coding},
  author={He, Dailan and Yang, Ziming and Peng, Weikun and Ma, Rui and Qin, Hongwei and Wang, Yan},
  booktitle={Proceedings of the IEEE/CVF conference on computer vision and pattern recognition},
  pages={5718--5727},
  year={2022}
}

@inproceedings{lic-attention,
  title={Learned image compression with discretized gaussian mixture likelihoods and attention modules},
  author={Cheng, Zhengxue and Sun, Heming and Takeuchi, Masaru and Katto, Jiro},
  booktitle={Proceedings of the IEEE/CVF conference on computer vision and pattern recognition},
  pages={7939--7948},
  year={2020}
}

@inproceedings{lic-balle2018variational,
  title={Variational image compression with a scale hyperprior},
  author={Ball{\'e}, Johannes and Minnen, David and Singh, Saurabh and Hwang, Sung Jin and Johnston, Nick},
  booktitle={International Conference on Learning Representations (ICLR)},
  year={2018}
}

@inproceedings{lic-progdtd,
  title={{ProgDTD}: Progressive learned image compression with double-tail-drop training},
  author={Hojjat, Ali and Haberer, Janek and Landsiedel, Olaf},
  booktitle={Proceedings of the IEEE/CVF Conference on Computer Vision and Pattern Recognition},
  pages={1130--1139},
  year={2023}
}

@inproceedings{lic-dpict,
  title={DPICT: Deep progressive image compression using trit-planes},
  author={Lee, Jae-Han and Jeon, Seungmin and Choi, Kwang Pyo and Park, Youngo and Kim, Chang-Su},
  booktitle={Proceedings of the IEEE/CVF conference on computer vision and pattern recognition},
  pages={16113--16122},
  year={2022}
}

@inproceedings{lic-ctc,
  title={Context-based trit-plane coding for progressive image compression},
  author={Jeon, Seungmin and Choi, Kwang Pyo and Park, Youngo and Kim, Chang-Su},
  booktitle={Proceedings of the IEEE/CVF Conference on Computer Vision and Pattern Recognition},
  pages={14348--14357},
  year={2023}
}

@inproceedings{cgic,
  title={Once-for-All: Controllable Generative Image Compression with Dynamic Granularity Adaptation},
  author={Li, Anqi and Li, Feng and Liu, Yuxi and Cong, Runmin and Zhao, Yao and Bai, Huihui},
  booktitle={International Conference on Learning Representations (ICLR)},
  year      = {2025},
}

@inproceedings{ms-illm,
  title={Improving statistical fidelity for neural image compression with implicit local likelihood models},
  author={Muckley, Matthew J and El-Nouby, Alaaeldin and Ullrich, Karen and J{\'e}gou, Herv{\'e} and Verbeek, Jakob},
  booktitle={International Conference on Machine Learning (ICML)},
  pages={25426--25443},
  year={2023},
  publisher =    {PMLR},
}

@article{hific,
  title={High-fidelity generative image compression},
  author={Mentzer, Fabian and Toderici, George D and Tschannen, Michael and Agustsson, Eirikur},
  journal={Advances in neural information processing systems},
  volume={33},
  pages={11913--11924},
  year={2020}
}

@article{ddpm,
  title={Denoising diffusion probabilistic models},
  author={Ho, Jonathan and Jain, Ajay and Abbeel, Pieter},
  journal={Advances in neural information processing systems},
  volume={33},
  pages={6840--6851},
  year={2020}
}

@inproceedings{stable-diffusion,
  title={High-resolution image synthesis with latent diffusion models},
  author={Rombach, Robin and Blattmann, Andreas and Lorenz, Dominik and Esser, Patrick and Ommer, Bj{\"o}rn},
  booktitle={Proceedings of the IEEE/CVF conference on computer vision and pattern recognition},
  pages={10684--10695},
  year={2022}
}

@inproceedings{diffusion-StableCodec,
  title={ StableCodec: Taming One-Step Diffusion for Extreme Image Compression},
  author={Tianyu, Zhang and Xin, Luo and Li, Li and Dong, Liu},
  booktitle = {International Conference on Computer Vision (ICCV)},
  year={2025}
}

@inproceedings{oscar,
  title={OSCAR: One-Step Diffusion Codec Across Multiple Bit-rates},
  author={Guo, Jinpei and Ji, Yifei and Chen, Zheng and Liu, Kai and Liu, Min and Rao, Wang and Li, Wenbo and Guo, Yong and Zhang, Yulun},
  booktitle={Conference on Neural Information Processing Systems (NeurIPS)},
  year      = {2025},
}

@inproceedings{diffusion-xue2025one,
  title={One-Step Diffusion-Based Image Compression with Semantic Distillation},
  author={Xue, Naifu and Jia, Zhaoyang and Li, Jiahao and Li, Bin and Zhang, Yuan and Lu, Yan},
  booktitle={Conference on Neural Information Processing Systems (NeurIPS)},
  year      = {2025},
}

@article{diffeic,
  title={Towards extreme image compression with latent feature guidance and diffusion prior},
  author={Li, Zhiyuan and Zhou, Yanhui and Wei, Hao and Ge, Chenyang and Jiang, Jingwen},
  journal={IEEE Transactions on Circuits and Systems for Video Technology},
  year={2025},
  volume  = {35},
  number  = {1},
  pages={888--899},
  doi     = {10.1109/TCSVT.2024.3455576},
  publisher={IEEE}
}

@article{diffusion-cdc,
  title={Lossy image compression with conditional diffusion models},
  author={Yang, Ruihan and Mandt, Stephan},
  journal={Advances in Neural Information Processing Systems},
  volume={36},
  pages={64971--64995},
  year={2023}
}

@inproceedings{diffusion-perco,
  title={Towards image compression with perfect realism at ultra-low bitrates},
  author={Careil, Marlene and Muckley, Matthew J and Verbeek, Jakob and Lathuili{\`e}re, St{\'e}phane},
  booktitle={International Conference on Learning Representations (ICLR)},
  year={2023}
}

@article{vq-1,
  title = {Image coding using vector quantization in the transform domain},
  journal = {Pattern Recognition Letters},
  volume = {1},
  number = {5},
  pages = {323-329},
  year = {1983},
  issn = {0167-8655},
  doi = {https://doi.org/10.1016/0167-8655(83)90071-5},
  url = {https://www.sciencedirect.com/science/article/pii/0167865583900715},
  author = {R.A. King and N.M. Nasrabadi}
}

@article{gan,
  title={Generative adversarial nets},
  author={Goodfellow, Ian J and Pouget-Abadie, Jean and Mirza, Mehdi and Xu, Bing and Warde-Farley, David and Ozair, Sherjil and Courville, Aaron and Bengio, Yoshua},
  journal={Advances in neural information processing systems},
  volume={27},
  year={2014}
}

@inproceedings{vq-gan,
  title={Taming transformers for high-resolution image synthesis},
  author={Esser, Patrick and Rombach, Robin and Ommer, Bjorn},
  booktitle={Proceedings of the IEEE/CVF conference on computer vision and pattern recognition},
  pages={12873--12883},
  year={2021}
}

@article{vq-vae,
  title={Neural discrete representation learning},
  author={Van Den Oord, Aaron and Vinyals, Oriol and others},
  journal={Advances in neural information processing systems},
  volume={30},
  year={2017}
}

@article{vq-vae-2,
  title={Generating diverse high-fidelity images with vq-vae-2},
  author={Razavi, Ali and Van den Oord, Aaron and Vinyals, Oriol},
  journal={Advances in neural information processing systems},
  volume={32},
  year={2019}
}

@inproceedings{vq-DLF,
  title={DLF: Extreme Image Compression with Dual-generative Latent Fusion},
  author={Xue, Naifu and Jia, Zhaoyang and Li, Jiahao and Li, Bin and Zhang, Yuan and Lu, Yan},
  booktitle = {International Conference on Computer Vision (ICCV)},
  year={2025}
}

@article{vq-gic,
  title={Generative latent coding for ultra-low bitrate image and video compression},
  author={Qi, Linfeng and Jia, Zhaoyang and Li, Jiahao and Li, Bin and Li, Houqiang and Lu, Yan},
  journal={IEEE Transactions on Circuits and Systems for Video Technology},
  year={2025},
  volume  = {35},
  number  = {10},
  pages={10500--10515},
  doi     = {10.1109/TCSVT.2025.3571944},
  publisher={IEEE}
}

@inproceedings{vq-lattice,
  title={Multirate Neural Image Compression with Adaptive Lattice Vector Quantization},
  author={Xu, Hao and Wu, Xiaolin and Zhang, Xi},
  booktitle={Proceedings of the Computer Vision and Pattern Recognition Conference},
  pages={7633--7642},
  year={2025}
}

@inproceedings{vq-zhu2022unified,
  title={Unified multivariate gaussian mixture for efficient neural image compression},
  author={Zhu, Xiaosu and Song, Jingkuan and Gao, Lianli and Zheng, Feng and Shen, Heng Tao},
  booktitle={Proceedings of the IEEE/CVF Conference on Computer Vision and Pattern Recognition},
  pages={17612--17621},
  year={2022}
}

@inproceedings{vq-feng2023nvtc,
  title={Nvtc: Nonlinear vector transform coding},
  author={Feng, Runsen and Guo, Zongyu and Li, Weiping and Chen, Zhibo},
  booktitle={Proceedings of the IEEE/CVF Conference on Computer Vision and Pattern Recognition},
  pages={6101--6110},
  year={2023}
}

@inproceedings{vq-mao2024extreme,
  title={Extreme image compression using fine-tuned {VQGANs}},
  author={Mao, Qi and Yang, Tinghan and Zhang, Yinuo and Wang, Zijian and Wang, Meng and Wang, Shiqi and Jin, Libiao and Ma, Siwei},
  booktitle={2024 Data Compression Conference (DCC)},
  pages={203--212},
  year={2024},
  organization={IEEE}
}

@article{llm-li2024misc,
  title={Misc: Ultra-low bitrate image semantic compression driven by large multimodal model},
  author={Li, Chunyi and Lu, Guo and Feng, Donghui and Wu, Haoning and Zhang, Zicheng and Liu, Xiaohong and Zhai, Guangtao and Lin, Weisi and Zhang, Wenjun},
  journal={IEEE Transactions on Image Processing},
  year={2024},
  volume  = {34},
  pages={335--349},
  doi     = {10.1109/TIP.2024.3515874},
  publisher={IEEE}
}

@article{llm-gao2025exploring,
  title={Exploring Multimodal Knowledge for Image Compression via Large Foundation Models},
  author={Gao, Junlong and Huang, Zhimeng and Mao, Qi and Ma, Siwei and Jia, Chuanmin},
  journal={IEEE Transactions on Image Processing},
  year={2025},
  volume  = {34},
  pages={5904--5919},
  doi     = {10.1109/TIP.2025.3607616},
  publisher={IEEE}
}

@inproceedings{lic-poelic,
  title={Po-elic: Perception-oriented efficient learned image coding},
  author={He, Dailan and Yang, Ziming and Yu, Hongjiu and Xu, Tongda and Luo, Jixiang and Chen, Yuan and Gao, Chenjian and Shi, Xinjie and Qin, Hongwei and Wang, Yan},
  booktitle={Proceedings of the IEEE/CVF Conference on Computer Vision and Pattern Recognition},
  pages={1764--1769},
  year={2022}
}

@InProceedings{perception,
  title = 	 {Rethinking Lossy Compression: The Rate-Distortion-Perception Tradeoff},
  author =       {Blau, Yochai and Michaeli, Tomer},
  booktitle = 	 {Proceedings of the 36th International Conference on Machine Learning},
  pages = 	 {675--685},
  year = 	 {2019},
  editor = 	 {Chaudhuri, Kamalika and Salakhutdinov, Ruslan},
  volume = 	 {97},
  series = 	 {Proceedings of Machine Learning Research},
  month = 	 {09--15 Jun},
  publisher =    {PMLR},
}

@inproceedings{lpips,
  title={The unreasonable effectiveness of deep features as a perceptual metric},
  author={Zhang, Richard and Isola, Phillip and Efros, Alexei A and Shechtman, Eli and Wang, Oliver},
  booktitle={Proceedings of the IEEE conference on computer vision and pattern recognition},
  pages={586--595},
  year={2018}
}

@article{dists,
  title={Image quality assessment: Unifying structure and texture similarity},
  author={Ding, Keyan and Ma, Kede and Wang, Shiqi and Simoncelli, Eero P},
  journal={IEEE transactions on pattern analysis and machine intelligence},
  volume={44},
  number={5},
  pages={2567--2581},
  year={2020},
  publisher={IEEE}
}

@article{speech-soundstream,
  title={Soundstream: An end-to-end neural audio codec},
  author={Zeghidour, Neil and Luebs, Alejandro and Omran, Ahmed and Skoglund, Jan and Tagliasacchi, Marco},
  journal={IEEE/ACM Transactions on Audio, Speech, and Language Processing},
  volume={30},
  pages={495--507},
  year={2021},
  publisher={IEEE}
}

@article{speech-dac,
  title={High-fidelity audio compression with improved rvqgan},
  author={Kumar, Rithesh and Seetharaman, Prem and Luebs, Alejandro and Kumar, Ishaan and Kumar, Kundan},
  journal={Advances in Neural Information Processing Systems},
  volume={36},
  pages={27980--27993},
  year={2023}
}

@inproceedings{adam,
  title={A method for stochastic optimization},
  author={Diederik, P., Kingma and Jimmy, Ba},
  booktitle={International Conference on Learning Representations (ICLR)},
  year={2015}
}

@article{fu2023vector,
  title={Vector quantized semantic communication system},
  author={Fu, Qifan and Xie, Huiqiang and Qin, Zhijin and Slabaugh, Gregory and Tao, Xiaoming},
  journal={IEEE Wireless Communications Letters},
  volume={12},
  number={6},
  pages={982--986},
  year={2023},
  publisher={IEEE}
}

@inproceedings{gic-1,
  title={Generative adversarial networks for extreme learned image compression},
  author={Eirikur, Agustsson and Michael, Tschannen and Fabian, Mentzer and Radu, Timofte and Luc, Van Gool},
  booktitle={Proceedings of the IEEE/CVF International Conference on Computer Vision},
  pages={221--231},
  year={2019}
}

@ARTICLE{trans,
  author={Goyal, V.K.},
  journal={IEEE Signal Processing Magazine}, 
  title={Theoretical foundations of transform coding}, 
  year={2001},
  volume={18},
  number={5},
  pages={9-21},
  doi={10.1109/79.952802}
}

@article{vae,
  title={Auto-encoding variational bayes},
  author={Kingma, Diederik P and Welling, Max},
  journal={arXiv preprint arXiv:1312.6114},
  year={2013}
}

@article{vgg,
  title={Very deep convolutional networks for large-scale image recognition},
  author={Simonyan, Karen and Zisserman, Andrew},
  journal={arXiv preprint arXiv:1409.1556},
  year={2014}
}

@inproceedings{imagenet,
  title={Imagenet: A large-scale hierarchical image database},
  author={Deng, Jia and Dong, Wei and Socher, Richard and Li, Li-Jia and Li, Kai and Fei-Fei, Li},
  booktitle={2009 IEEE conference on computer vision and pattern recognition (CVPR)},
  pages={248--255},
  year={2009},
  organization={IEEE}
}

@inproceedings{ms-ssim,
  title={Multiscale structural similarity for image quality assessment},
  author={Wang, Zhou and Simoncelli, Eero P and Bovik, Alan C},
  booktitle={The thirty-seventh asilomar conference on signals, systems \& computers, 2003},
  volume={2},
  pages={1398--1402},
  year={2003},
  organization={IEEE}
}

@inproceedings{relu,
  title={Rectified linear units improve restricted boltzmann machines},
  author={Nair, Vinod and Hinton, Geoffrey E},
  booktitle={Proceedings of the 27th international conference on machine learning (ICML-10)},
  pages={807--814},
  year={2010}
}

@article{satellite,
  author={Kodheli, Oltjon and Lagunas, Eva and Maturo, Nicola and Sharma, Shree Krishna and Shankar, Bhavani and Montoya, Jesus Fabian Mendoza and Duncan, Juan Carlos Merlano and Spano, Danilo and Chatzinotas, Symeon and Kisseleff, Steven and Querol, Jorge and Lei, Lei and Vu, Thang X. and Goussetis, George},
  journal={IEEE Communications Surveys \& Tutorials}, 
  title={Satellite Communications in the New Space Era: A Survey and Future Challenges}, 
  year={2021},
  volume={23},
  number={1},
  pages={70-109},
  doi={10.1109/COMST.2020.3028247}
}

@inproceedings{range-coding,
  title={Range encoding: an algorithm for removing redundancy from a digitised message},
  author={Martin, G Nigel N},
  booktitle={Proc. Institution of Electronic and Radio Engineers International Conference on Video and Data Recording},
  volume={2},
  year={1979}
}

@article{bd-rate,
  title={Calculation of average PSNR differences between RD-curves},
  author={Bjontegaard, Gisle},
  journal={ITU-T SG16, Doc. VCEG-M33},
  year={2001}
}

@inproceedings{clip-iqa,
  title={Exploring clip for assessing the look and feel of images},
  author={Wang, Jianyi and Chan, Kelvin CK and Loy, Chen Change},
  booktitle={Proceedings of the AAAI conference on artificial intelligence},
  volume={37},
  number={2},
  pages={2555--2563},
  year={2023}
}

@inproceedings{diffusion-prog,
  title={Progressive compression with universally quantized diffusion models},
  author={Yang, Yibo and Will, Justus C and Mandt, Stephan},
  booktitle={International Conference on Learning Representations (ICLR)},
  year={2025}
}
}
\clearpage
\setcounter{page}{1}
\maketitlesupplementary

\renewcommand{\thesection}{\Alph{section}}
\setcounter{section}{0}

\section{Progressive Decoding}

Many GICs are designed for extremely low-bitrate scenarios~\cite{vq-gic,vq-DLF,diffusion-StableCodec,diffusion-xue2025one}, where progressive decoding plays a critical role. Progressive decoding enables a usable image preview from only a portion of the bitstream, without waiting for the full transmission to complete, thus allowing the receiver to react faster to the image content. 
This requirement is especially prevalent in satellite communications~\cite{satellite}.
A concrete usage example in a satellite communication setting is provided later in this section.

\begin{figure}[t]
  \centering
  \includegraphics[width=0.95\linewidth]{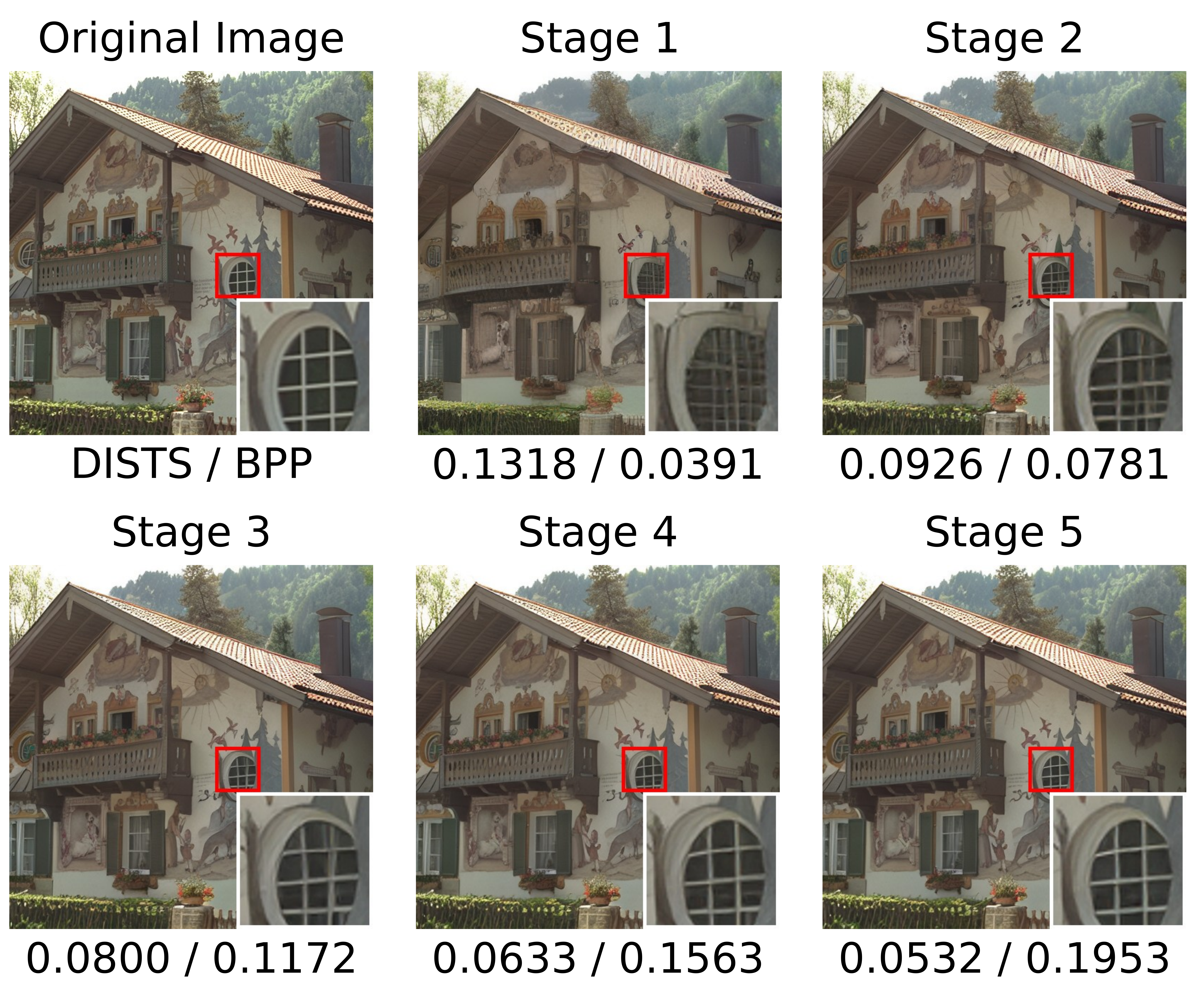}
  \caption{Visualization of progressive image transmission with ProGIC. Reconstructions are shown across increasing stages, where the cumulative BPP increases and visual fidelity improves. Values beneath each image denote DISTS / BPP.}
  \label{fig:prog}
\end{figure}

The progressive decoding behavior of ProGIC is illustrated in~\cref{fig:prog}. 
After decoding the first stage, the main semantic content of the image is already largely recovered, which meets the demand for rapid preview. 
As more bits from subsequent stages arrive, the reconstructed image gradually approaches the original, and the visual discrepancy diminishes.

\section{BPP Selection}

\begin{figure}[t]
  \centering
  \includegraphics[width=0.95\linewidth]{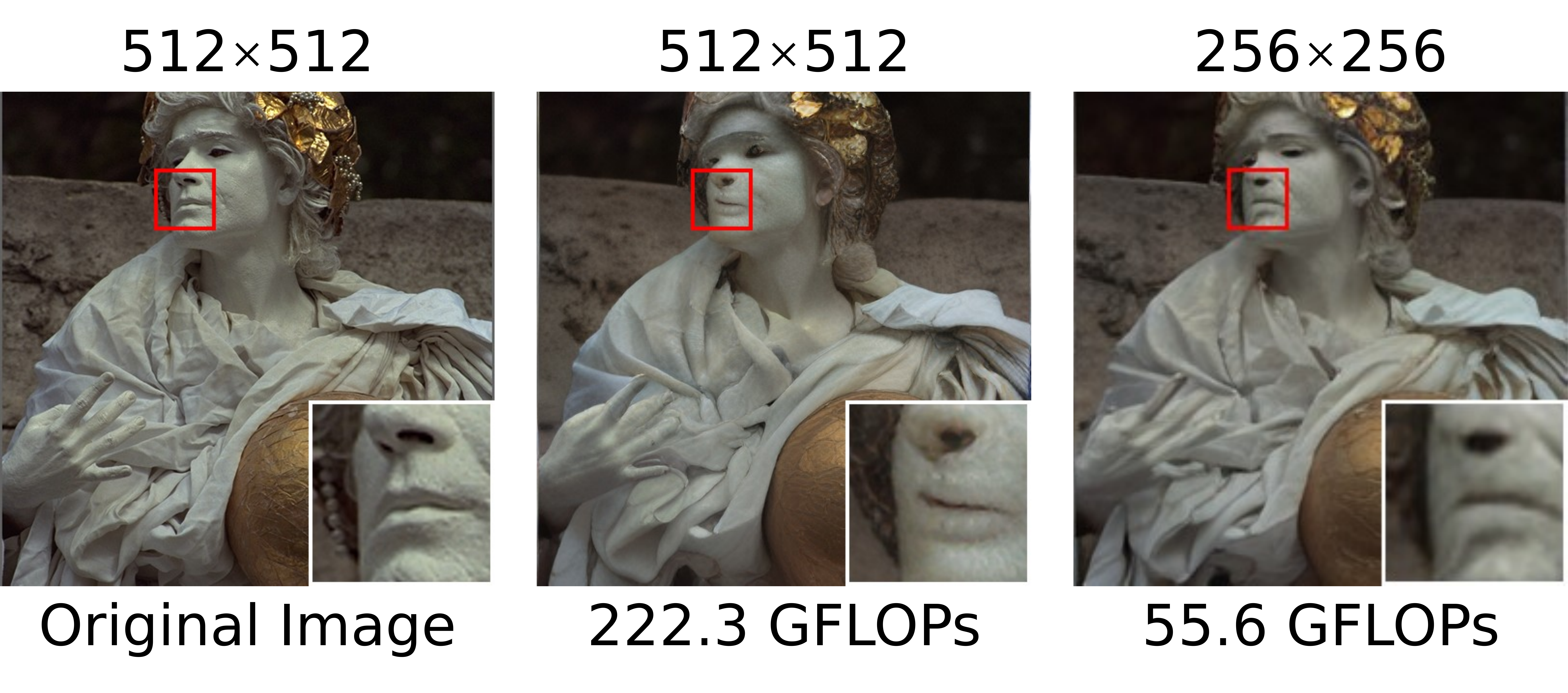}
  \caption{Reconstruction quality comparison across different resolutions, both compressed to a total of 1280 bytes. The left column shows the original image. The middle column shows the reconstructed image at 0.04 BPP and $512\times 512$ resolution. The right column shows the reconstructed image at 0.16 BPP and $256\times 256$ resolution. Resolution and compression FLOPs are indicated above and below each result, respectively.}
  \label{fig:percp}
\end{figure}

Unlike GICs designed for extremely low-bitrate scenarios~\cite{vq-gic,vq-DLF,diffusion-StableCodec,diffusion-xue2025one}, ProGIC targets higher bitrates. 
In practical transmission, latency is determined by the total bitstream length rather than by BPP. 
For example, a $256\times256$ image at 0.16 BPP and a $512\times512$ image at 0.04 BPP both yield approximately the same bitstream size of 1280 bytes. \cref{fig:percp} compares the reconstruction quality under these two configurations. 

We observe that at lower BPP, models tend to ``generate'' fine details, often producing unrealistic textures. 
In contrast, the lower resolution and higher BPP setting is slightly blurrier in overall appearance than the full resolution reconstruction, but it recovers image details more faithfully.
Such faithful detail preservation is crucial in low-bitrate applications such as emergency response and satellite communications, where accurate decisions rely on authentic visual cues. 

Moreover, using lower-resolution inputs leads to significantly fewer FLOPs, resulting in faster encoding, which is especially beneficial on edge devices. 
The middle image in~\cref{fig:percp} requires 222.3 GFLOPs to process, whereas the right image requires only 55.6 GFLOPs.
Nevertheless, the higher BPP configuration is preferred to ensure the authenticity and reliability of reconstructed details.

\section{Codebook Properties}

In this section, we further analyze the properties of the codebooks in ProGIC.

\subsection{Entropy of Codebook Indices}
\begin{figure}[t]
  \centering
  \includegraphics[width=0.90\linewidth]{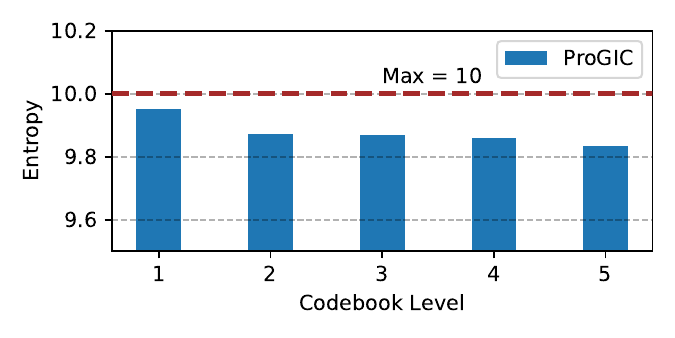}
  \caption{Entropy of different codebook usages in ProGIC.}
  \label{fig:codebook_entropy}
\end{figure}

As discussed in the main text, after RVQ the entropy of the codewords is very high, which results in limited gains from entropy coding. 
To verify this, all RVQ indices obtained on the ImageNet dataset are collected, their empirical probability distribution is computed, and the resulting entropy is measured, as shown in~\cref{fig:codebook_entropy}. 
Since each codebook contains $2^{10} = 1024$ codewords, the maximum possible entropy is 10, which corresponds to a uniform distribution where all indices occur with equal probability. 
Our results show that the entropy of every codebook exceeds 9.80, indicating that the codewords are used nearly uniformly.

\subsection{Results for Entropy Coding}
\begin{figure}[t]
  \centering
  \includegraphics[width=0.95\linewidth]{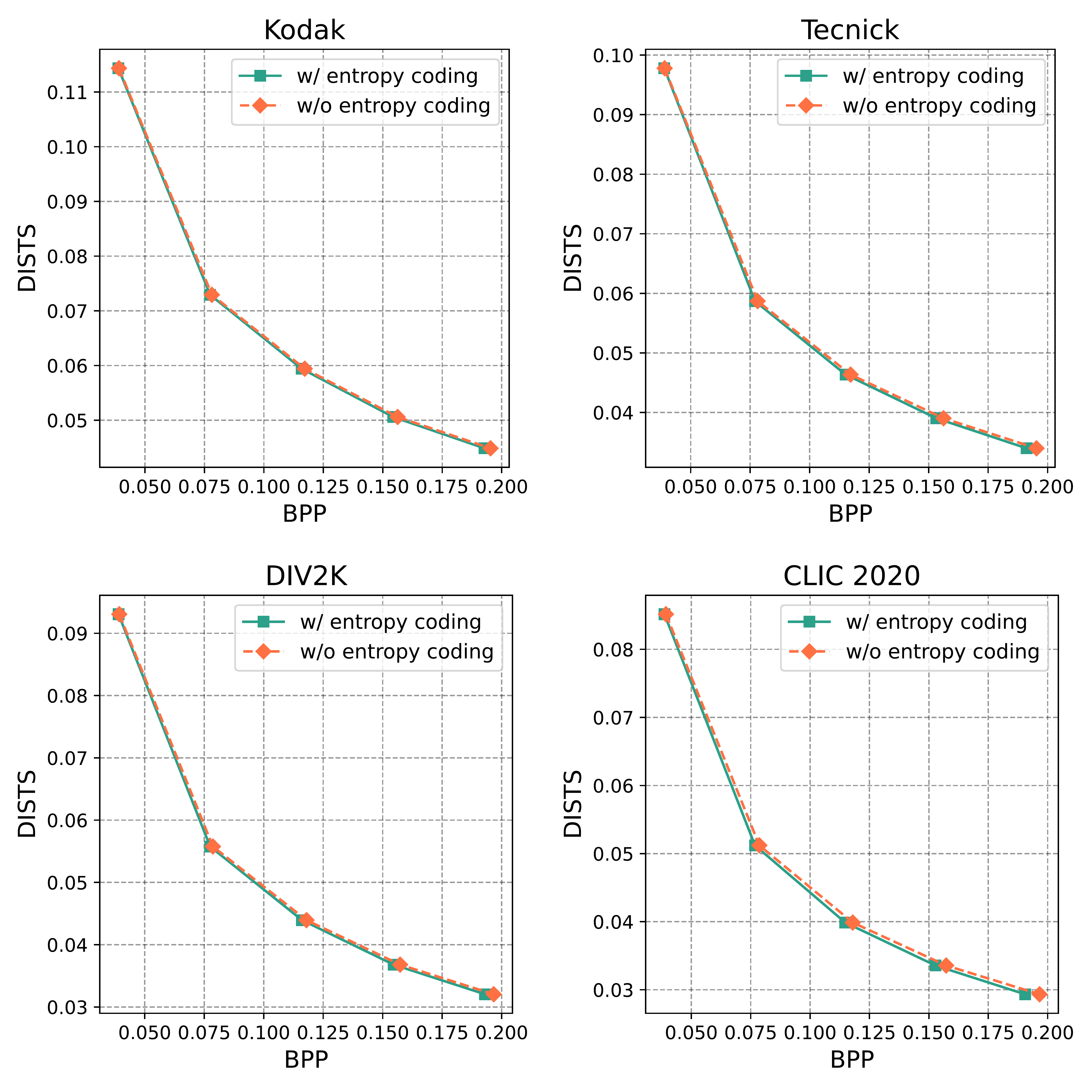}
  \caption{Rate-distortion curves with and without entropy coding.}
  \label{fig:entropy_coding_curve}
\end{figure}

As shown in~\cref{fig:entropy_coding_curve}, we compare rate-distortion (R--D) curves with and without entropy coding across different datasets. 
We adopt range coding~\cite{range-coding}, which is consistent with most existing learned image compression methods~\cite{lic-balle2018variational,lic-mlic,lalic,hpcm}. 
Since the prior probability distribution used by the entropy coder is learned from a large amount of data, there is an inevitable discrepancy between this distribution and the true probability distribution of image indices. 
As a result, the achieved rates are slightly above the theoretical entropy limit, which further reduces the practical gain from entropy coding. 
Considering all these factors, entropy coding is not used in ProGIC.

\subsection{Analysis for Latent Features}
\begin{figure}[t]
  \centering
  \includegraphics[width=0.85\linewidth]{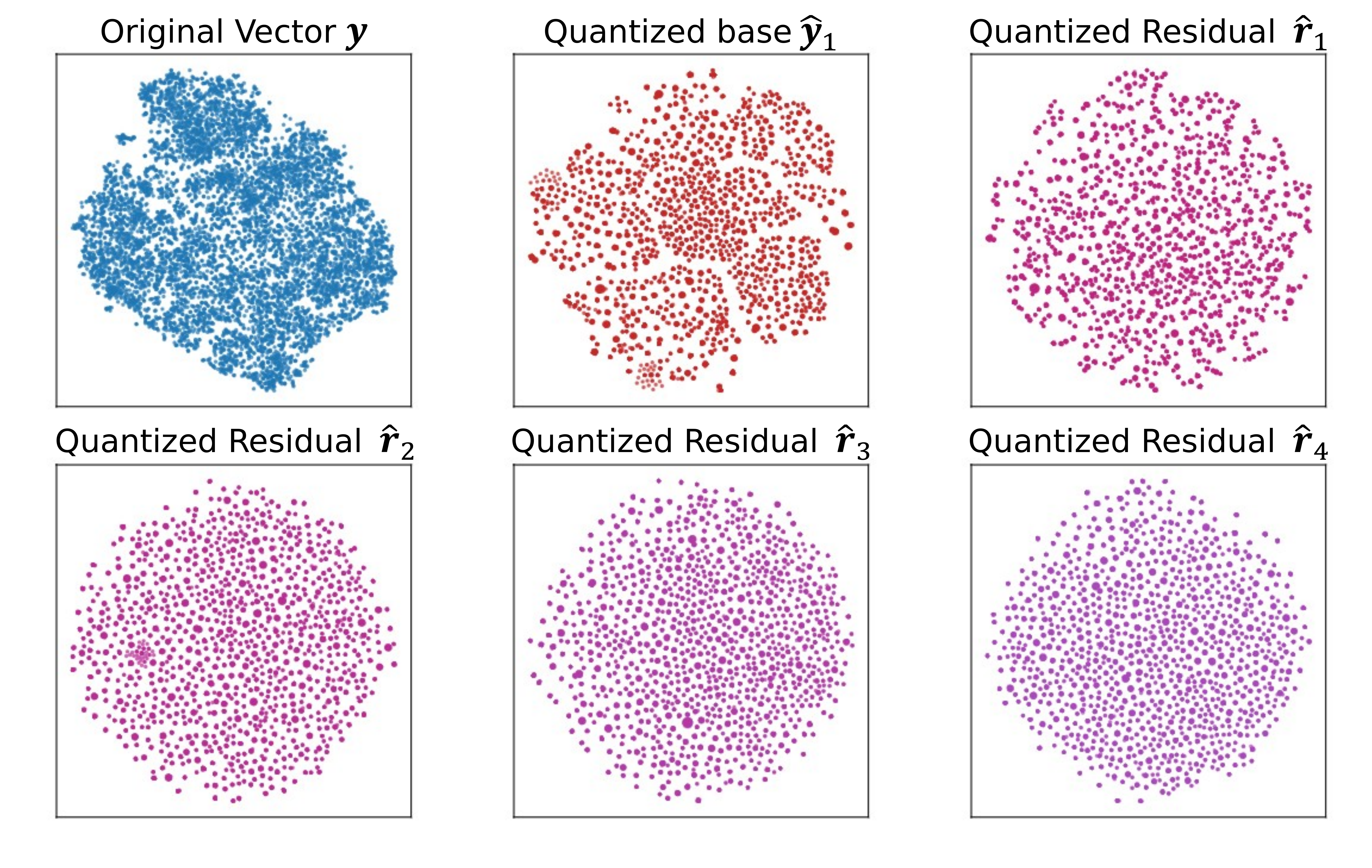}
  \caption{Visualization by t-SNE of latent features. The top-left plot shows the distribution of the unquantized latent vector $\boldsymbol{y}$. The subsequent plots illustrate the distributions of the base vector $\hat{\boldsymbol{y}}_1$ and the residuals from each quantization stage, respectively.}
  \label{fig:codebook_view}
\end{figure}

Furthermore, we analyze the latent features $\boldsymbol{y}$ and the quantized residuals at each stage on the Kodak~\cite{Kodak} dataset, and visualize them using t-SNE, as shown in~\cref{fig:codebook_view}. 
The original latent features $\boldsymbol{y}$ exhibit a relatively concentrated distribution. After the first-stage quantization, the quantized vectors show mild dispersion. In contrast, the residuals from subsequent stages are distributed more uniformly across the space. 
This suggests that early codebooks capture high-level semantic information, whereas later codebooks focus on fine-grained differences between the reconstructed and original features, leading to more uniform spatial distributions in their visualizations.
This observation aligns with the idea of DLF~\cite{vq-DLF}, which directly models residuals using an entropy model and entropy coding. 
Similarly, ProGIC employs multiple additional codebooks to represent these residuals.

\begin{figure*}[t]
  \centering
  \begin{subfigure}[c]{0.42\linewidth}
    \includegraphics[width=0.9999\linewidth]{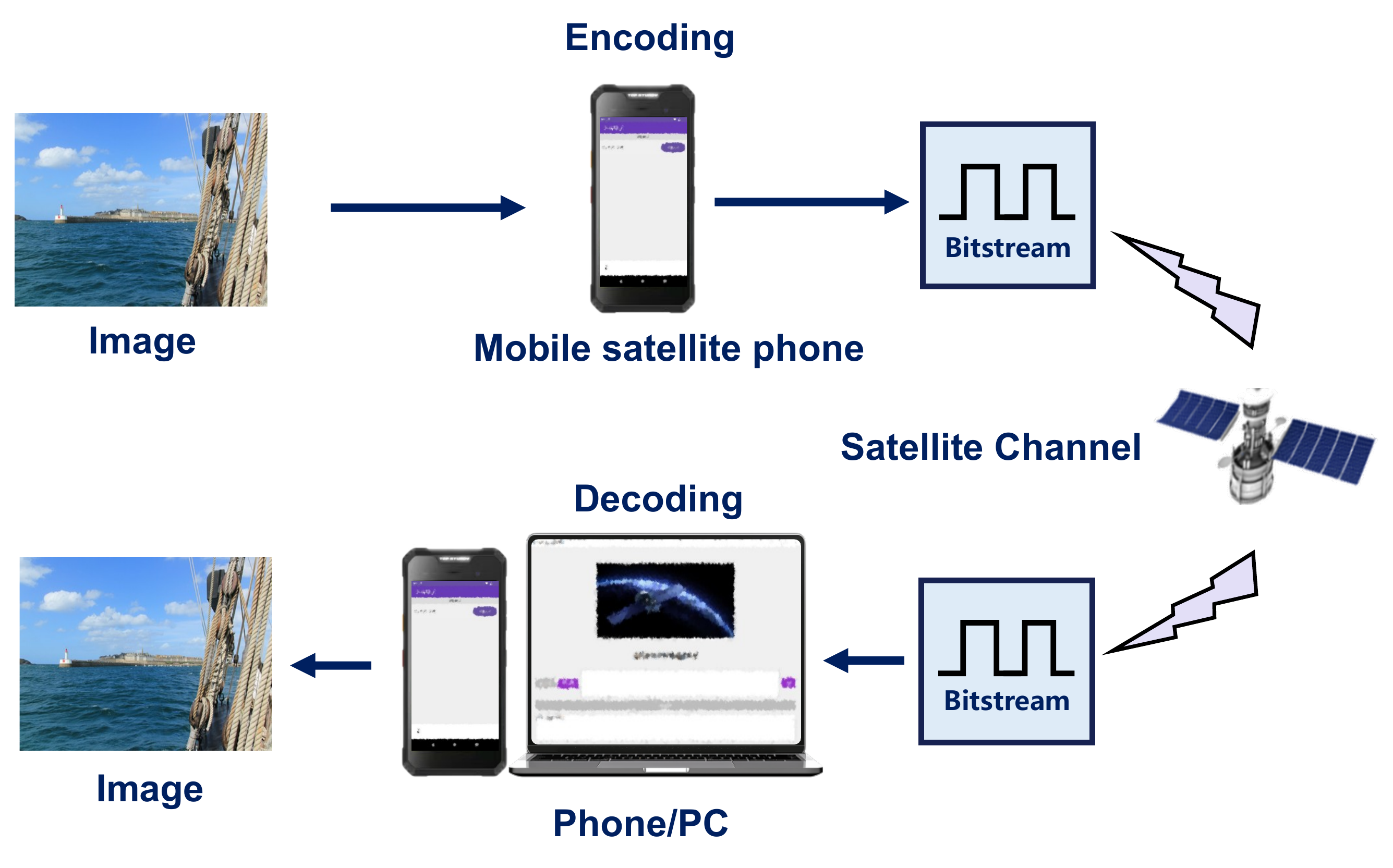}
    \caption{Satellite communication for mobile phone deployment.}
    \label{fig:satellite}
  \end{subfigure}
  \hfill
  \begin{subfigure}[c]{0.57\linewidth}
    \includegraphics[width=0.9999\linewidth]{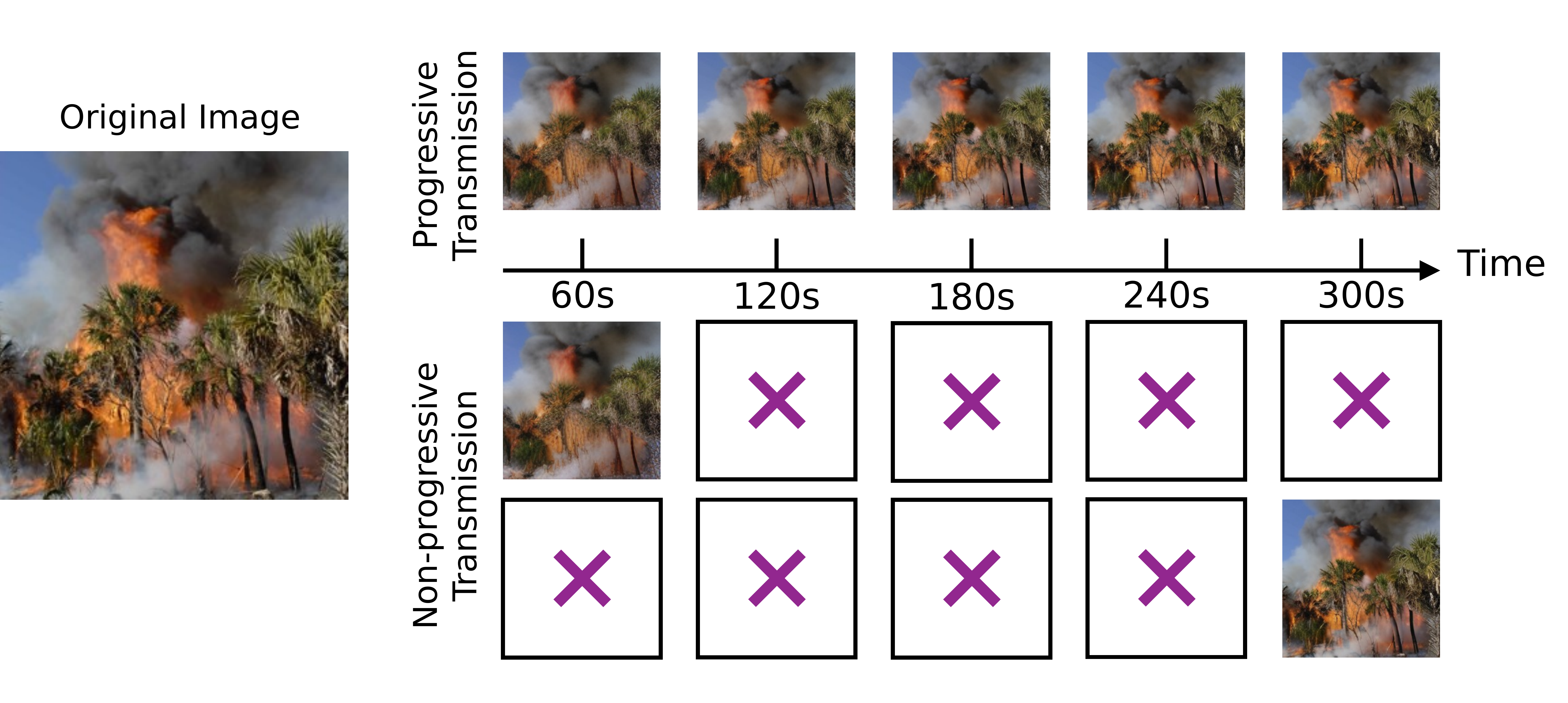}
    \caption{Comparison between progressive and non-progressive transmission.}
    \label{fig:forest_fire}
  \end{subfigure}
  \caption{In the event of a forest fire, ProGIC enables rapid response by transmitting images over a satellite short message link, assuming one transmission every 60 seconds. The leftmost column shows the original image. The first row illustrates progressive transmission: a usable preview is available immediately and is progressively refined as more bits arrive. In contrast, the second and third rows depict non-progressive schemes, which require the full bitstream before decoding. The second row uses a low BPP, yielding a fast but fixed-quality preview that cannot be improved later. The third row uses a high BPP, delivering higher fidelity but only after a long wait for the complete transmission, which delays situational awareness.}
  \label{fig:sallite_fire}
\end{figure*}

\subsection{More BPP Granularities with different codebook untilizations}

In the main text, results are reported for models at BPP values of ${0.0391, 0.0781, 0.1172, 0.1562, 0.1953}$. 
The achievable BPP range and step size are determined by three key parameters: the number of codebooks $N$, the number of codewords per codebook $2^L$, and the downsampling factor $f$, according to
\begin{equation}
  \text{BPP} = \frac{L}{f \times f} \times N,
\end{equation}
where $\frac{L}{f \times f}$ represents the minimum BPP increment per progressive stage. 

\begin{figure}[t]
  \centering
  \includegraphics[width=0.90\linewidth]{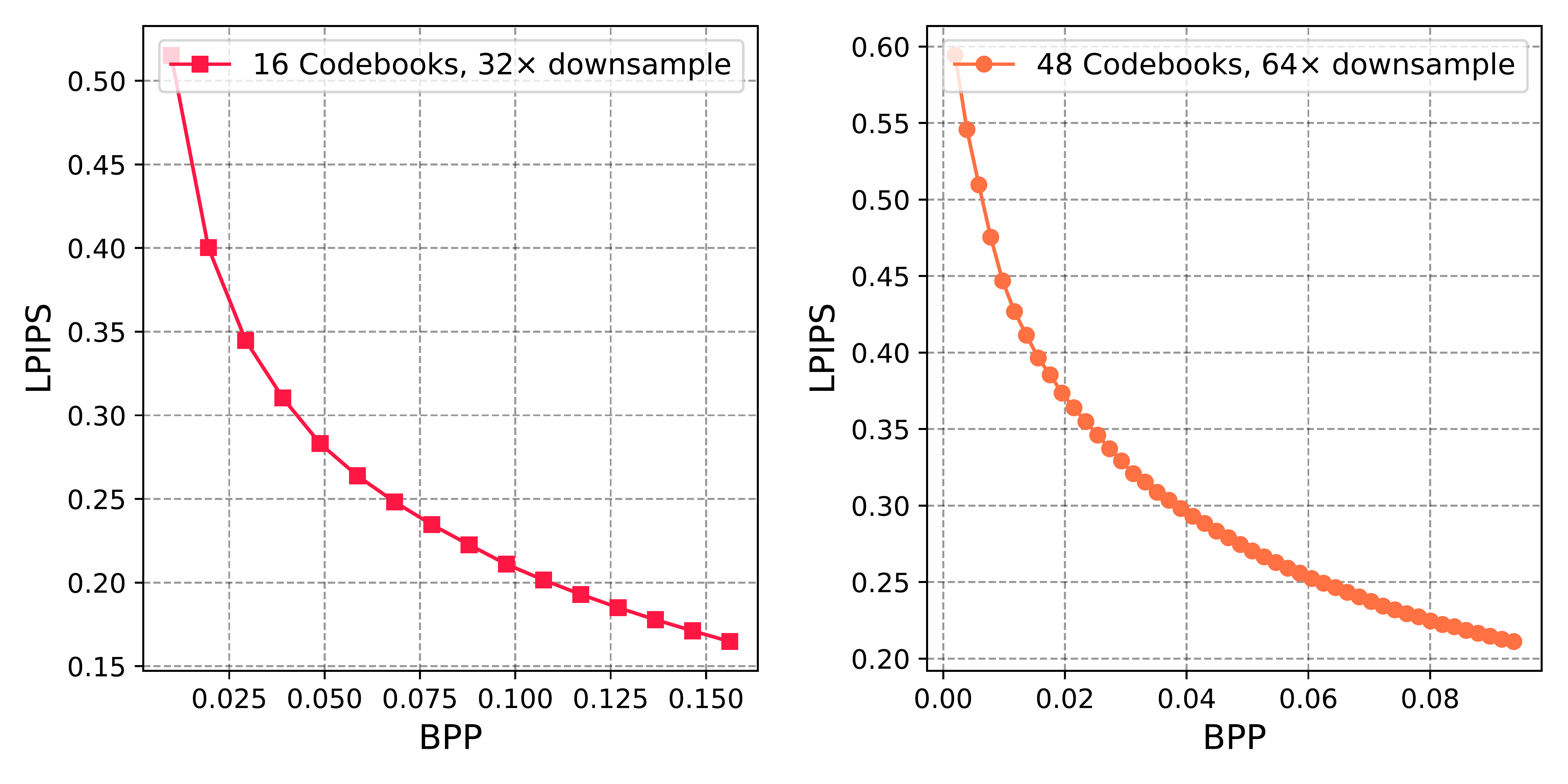}
  \caption{Different BPP ranges achieved by varying the number of codebooks and downsampling factors. The left panel shows the case of 16 codebooks with a downsampling factor of 32, with a minimal BPP increment of 0.009766. The right panel shows the case of 48 codebooks with a downsampling factor of 64, with a minimal BPP increment of 0.001953.}
  \label{fig:down}
\end{figure}

As shown in~\cref{fig:down}, different combinations of $N$ and $f$ yield different BPP ranges. 
The left panel corresponds to $N=16$, $L=10$, and $f=32$, supporting 16 progressive stages with a minimal BPP increment of $0.009766$. 
The right panel shows $N=48$, $L=8$, and $f=64$, enabling 48 stages with a finer BPP increment of $0.001953$. 
By adjusting these parameters, ProGIC can flexibly accommodate diverse transmission requirements.

\section{Use Case: Mobile Phone Deployment for Satellite Communication}

In this section, a practical application of ProGIC in satellite short message communication is presented—a critical solution for remote areas that lack cellular coverage, where satellite links are often the only means of connectivity. Timely disaster monitoring in such regions is vital, as early detection and rapid response can significantly mitigate losses. 
For instance, in the case of a forest fire, prompt identification enables immediate containment measures to prevent large-scale spread.

The satellite communication system is illustrated in~\cref{fig:satellite}. 
Existing satellite short message systems typically support payloads of a few hundred bytes per packet, with transmission intervals ranging from 10 to 60 seconds. 
Assuming a payload capacity of 360 bytes per packet with a 60-second interval, ProGIC is configured with $N=5$, $L=10$, and $f=16$, so that each packet carries 320 bytes of compressed data. The ProGIC model is exported to TorchScript format, packaged as an SDK, and deployed on a satellite short message handheld device powered by the MediaTek Helio P35 SoC for on-device encoding and decoding. A Java-based web interface is also developed for decoding and visualization on desktop clients. Because ProGIC does not use entropy coding, it avoids the cross-platform precision inconsistencies commonly observed in other learned image codecs~\cite{dcvc-rt}, ensuring reliable and consistent decoding across diverse hardware platforms.

The simulation results are shown in~\cref{fig:forest_fire}. 
Owing to its progressive design, ProGIC delivers an immediate coarse preview upon receiving the first packet, and the image quality progressively improves as subsequent packets arrive. 
In contrast, non-progressive codecs must wait for all packets to be received before decoding. 
If the BPP is set too low, the final reconstruction lacks clarity. If it is set too high, the waiting time becomes impractical. 
This demonstrates that the progressive transmission of ProGIC is particularly well suited to bandwidth- and latency-constrained satellite short message scenarios.

\section{Additional Experimental Results}

\subsection{Performance Details}

This section provides additional details regarding the baselines mentioned in the main text.

For VTM~\cite{vtm}, its intra-frame coding mode is currently one of the strongest image compression methods available. We use the relatively recent VTM-23.10 version in our experiments to ensure a fair comparison in terms of encoding and decoding time. We build the VTM project on a Linux system and run tests with the following command:

\begin{verbatim}
EncoderApp
  -i [input.yuv]
  -c encoder_intra_vtm.cfg
  -o [output.yuv] 
  -b [output.bin]
  --wdt [width] 
  --hgt [height] 
  -q [QP]
  --InputBitDepth=8 
  -fr 1 
  -f 1
  --InputChromaFormat=420
\end{verbatim}

We use YUV420-formatted input images, as this configuration yields faster runtimes. Comparing speed under this optimized setup ensures a fair evaluation.

For Control-GIC~\cite{cgic}, we perform an exhaustive search over all granularity combinations with a step size of $0.01$ and retain the best-performing configuration. 
We observe that Control-GIC suffers significant performance degradation when the BPP falls below 0.15. 
To remain consistent with the original paper, we exclude results in the BPP $<$ 0.15 range when computing BD-rate. 
Additionally, we find that the encoding and decoding complexity of Control-GIC scales quadratically with the number of image pixels, whereas other models scale linearly. 
At a resolution of $256 \times 256$, our measured encoding and decoding times closely match those reported in the original paper. 
However, at the standard Kodak resolution of $512 \times 768$, Control-GIC becomes significantly slower. On DIV2K and CLIC2020, we use the official tiling function to avoid out-of-memory errors.

For OSCAR~\cite{oscar}, we use the author-provided pretrained models and code for evaluation. However, the official implementation does not support high-resolution image testing. To avoid introducing artificial bias, we retain the out-of-memory behavior on DIV2K and CLIC2020.

The official implementation of HiFiC~\cite{hific} relies on an older version of TensorFlow and is incompatible with modern GPUs such as the NVIDIA A100 or RTX 4090. 
For a fair comparison, we instead use a community-provided PyTorch reimplementation with its pretrained weights. All other models use official implementations and pretrained checkpoints.

We compute LPIPS~\cite{lpips} using the \texttt{lpips} Python library, normalizing inputs to $[-1, 1]$ as in the official setup and using the pretrained VGG~\cite{vgg} network weights, which correlate better with perceptual quality than AlexNet. 
DISTS~\cite{dists} is computed using the \texttt{DISTS\_pytorch} library with inputs normalized to $[0, 1]$, following the official configuration. 
FLOPs are measured using the \texttt{calflops} library in Python, where we adopt the convention that $1~\text{FLOP} = 2~\text{MACs}$.

BD-rate (Bjøntegaard Delta rate)~\cite{bd-rate} is a widely used metric to compare the average bitrate savings of one method over another across a range of quality levels. It computes the area between two rate--distortion (R--D) curves after interpolating them using a monotonic piecewise cubic Hermite interpolating polynomial (PCHIP). 
A negative BD-rate indicates that the proposed method achieves the same quality at a lower bitrate compared with the baseline. 
We use the \texttt{bjontegaard} Python library for these calculations.

\begin{table}[t]
  \caption{Comparison of BD-rate on the Kodak, Tecnick, DIV2K, and CLIC datasets measured with DISTS. Best results are in \textbf{bold}. Second-best are \underline{underlined}.}
  \label{tab:dists}
  \centering
  \begingroup
  \setlength{\tabcolsep}{3pt} 
  \begin{tabular}{@{}lrrrr@{}}
    \toprule
    \multirow{2}{*}{Method} &
    \multicolumn{4}{c}{BD-rate (DISTS)} \\
    \cmidrule(lr){2-5}
    & Kodak & Tecnick & DIV2K & CLIC\\
    \midrule
    HiFiC~\cite{hific} & 90.08\%   & 99.67\%   & 100.76\%   & 124.45\%  \\
    Control-GIC~\cite{cgic} & 34.18\%   & 67.12\%   & 62.09\%   & 110.76\%  \\
    MS-ILLM~\cite{ms-illm}  & 0.00\% & 0.00\% & 0.00\% & \underline{0.00\%} \\
    DiffEIC~\cite{diffeic}  & -33.79\%  & 23.68\%   & 15.78\%        & 59.91\%        \\
    OSCAR~\cite{oscar} & \underline{-50.63\%}  & \underline{-4.76\%}    & --       & --          \\
    \midrule
    ProGIC-s (Ours) & -43.87\%  & 2.86\%    & \underline{-30.62\%}  & 11.86\%  \\
    ProGIC (Ours)  & \textbf{-57.57\%}  & \textbf{-20.95\%}   & \textbf{-44.49\%}  & \textbf{-13.19\%}  \\

    \bottomrule
  \end{tabular}
  \endgroup
\end{table}

\subsection{Quantitative Results on DISTS}

In the main text, BD-rate results are reported based on LPIPS. 
In this section, we also provide BD-rate results based on DISTS, as shown in~\cref{tab:dists}. ProGIC achieves the best performance across all datasets.

\subsection{Runtime Analysis on High-Resolution Images}

In this section, we evaluate the encoding and decoding time and GPU memory usage of various models on an NVIDIA A100 across different resolutions. 
The results are shown in~\cref{tab:high_resolution_time}. 
Because ProGIC completes both encoding and decoding in under 200 ms at 4K resolution, any model taking longer than 10s is marked as $>10$s. 
For memory usage, models that encounter out-of-memory errors at high resolutions are marked as $>80$. 
These results demonstrate that ProGIC achieves significant speed and memory advantages over traditional codecs, non-generative codecs, and other GICs across all tested resolutions.

\begin{table*}[t]
  \caption{Comparison of GPU runtimes (ms) and memory (GB) for image encoding and decoding across different resolutions. Enc./Dec. denote encoding/decoding times. Mem. denotes memory usage. Best results are in \textbf{bold}. Second-best are \underline{underlined}.}
  \label{tab:high_resolution_time}
  \centering
  \begingroup
  \setlength{\tabcolsep}{3.2pt} 
  \begin{tabular}{@{}lrrrrrrrrrrrr@{}}
    \toprule
    \multirow{2}{*}{Method} & \multicolumn{3}{c}{512$\times$768} & \multicolumn{3}{c}{1080$\times$1920} & \multicolumn{3}{c}{1440$\times$2560} & \multicolumn{3}{c}{2160$\times$3840}\\
    \cmidrule(lr){2-4} \cmidrule(lr){5-7} \cmidrule(lr){8-10} \cmidrule(lr){11-13}
     & Enc. & Dec. & Mem. & Enc. & Dec. & Mem. & Enc. & Dec. & Mem. & Enc. & Dec. & Mem. \\
    \midrule
    VTM-23.10~\cite{vtm} & $>$10s & 150.30 & -- & $>$10s & 230.10 & -- & $>$10s & 288.16 & -- & $>$10s & 486.71 & --  \\
    LIC-HPCM~\cite{hpcm} & 62.37 & 82.88 & 0.53 & 309.80 & 342.95 & 1.98 & 465.91 & 474.49 & 2.84 & 1121.92 & 1147.79 & 6.05  \\
    DCVC-RT~\cite{dcvc-rt} & 14.09 & 17.08 & \underline{0.34} & 76.68 & 59.87 & \underline{1.04} & 135.87 & 102.95 & 1.73 & 259.86 & 197.83 & 3.63  \\
    \midrule
    HiFiC~\cite{hific} & 526.51  & 1408.60 & 1.14 & 2894.55 & 6909.92 & 2.97 & 5179.44 & $>$10s & 4.78 & $>$10s & $>$10s & 9.75  \\
    Control-GIC~\cite{cgic} & 103.56  & 436.26 & 6.53 & 610.76 & 2186.30 & 69.99 & 1190.97 & 4361.94 & 34.93 & -- & $>$10s & --  \\
    MS-ILLM~\cite{ms-illm} & 165.38  & 147.79 & 1.12 & 350.85 & 379.01 & 2.99 & 516.47 & 601.18 & 4.87 & 1305.93 & 1613.82 & 9.94  \\
    DiffEIC~\cite{diffeic} & 210.18 & 4661.74 & 6.86 & -- & $>$10s & -- & -- & $>$10s & -- & -- & $>$10s & --  \\
    OSCAR~\cite{oscar} & 53.04 & 167.56 & 5.57 & 513.20 & 1123.38 & 24.75 & -- & -- & $>$80 & -- & -- & $>$80 \\
    \midrule
    ProGIC (Ours) & \underline{7.64} & \underline{10.99} & 0.66 & \underline{28.40} & \underline{55.00} & 1.07 & \underline{38.63} & \underline{80.38} & \underline{1.46} & \underline{78.26} & \underline{145.24} & \underline{2.58}  \\
    ProGIC-s (Ours) & \textbf{6.13} & \textbf{7.66} & \textbf{0.34} & \textbf{10.32} & \textbf{17.24} & \textbf{0.66} & \textbf{15.81} & \textbf{29.07} & \textbf{0.95} & \textbf{35.22} & \textbf{63.05} & \textbf{1.79}  \\
    \bottomrule
    
  \end{tabular}
  \endgroup
\end{table*}

\subsection{Quantitative Results for Other Metrics}

\begin{figure*}[t]
  \centering
  \includegraphics[width=0.99\linewidth]{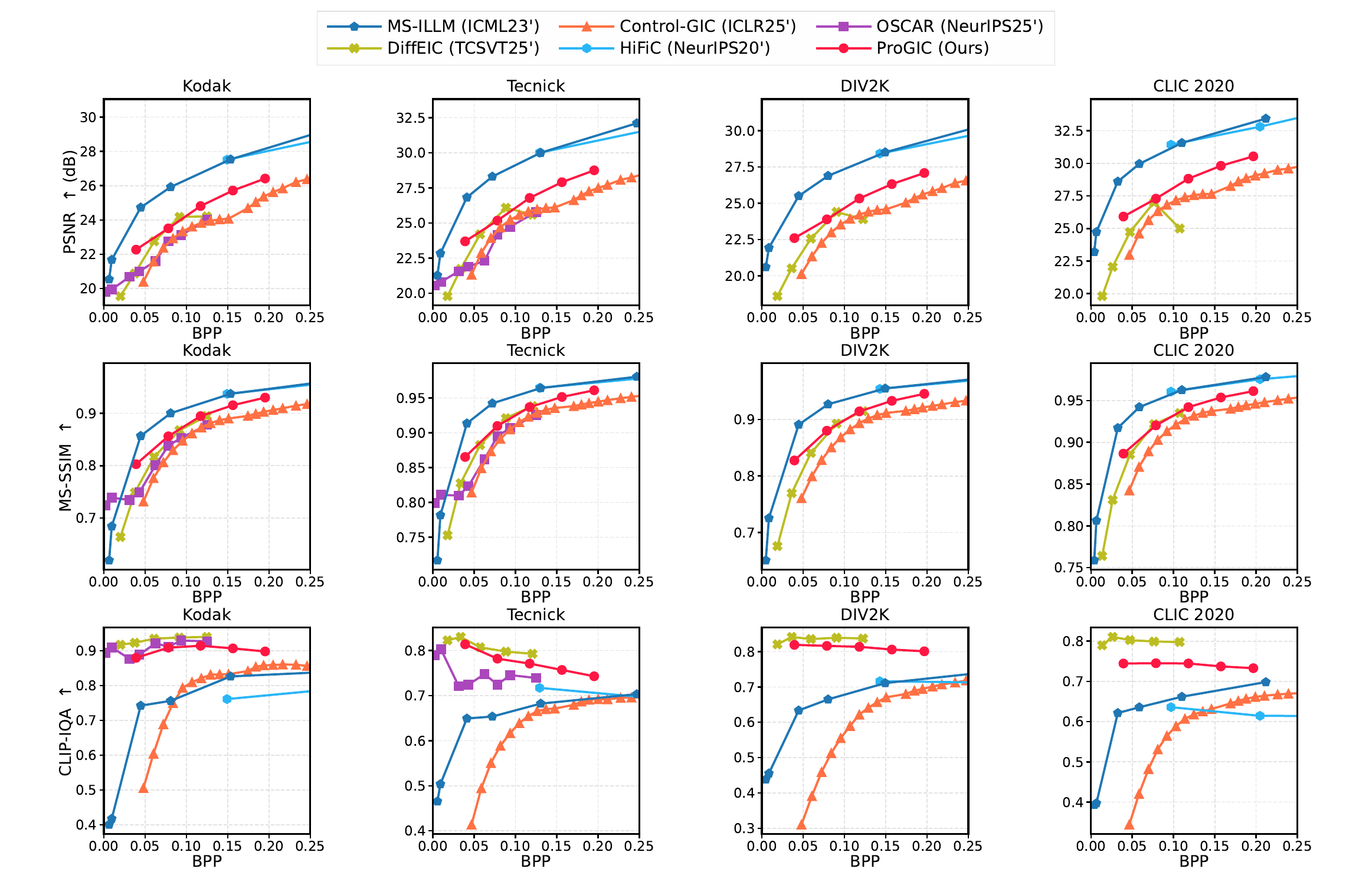}
  \caption{Rate-distortion performance on the Kodak, Tecnick, DIV2K, and CLIC2020-Professional datasets, evaluated with PSNR, MS-SSIM~\cite{ms-ssim}, and CLIP-IQA~\cite{clip-iqa} versus BPP. Curves closer to the lower left are better, indicating higher quality at the same compression ratio.}
  \label{fig:RD-curve-2}
\end{figure*}

In this section, we report R--D performance on the Kodak, Tecnick, DIV2K, and CLIC2020-Professional datasets, evaluated using PSNR, MS-SSIM~\cite{ms-ssim}, and CLIP-IQA~\cite{clip-iqa}, as shown in~\cref{fig:RD-curve-2}. ProGIC lags behind in pixel-level metrics such as PSNR because higher perceptual quality often comes at the cost of lower pixel-wise fidelity~\cite{perception}. While its CLIP-IQA score is slightly inferior to that of DiffEIC, DiffEIC is over two orders of magnitude slower in decoding, making it impractical for real-world applications.

\subsection{More Visualization Results}

\begin{figure*}[t]
  \centering
  \includegraphics[width=0.99\linewidth]{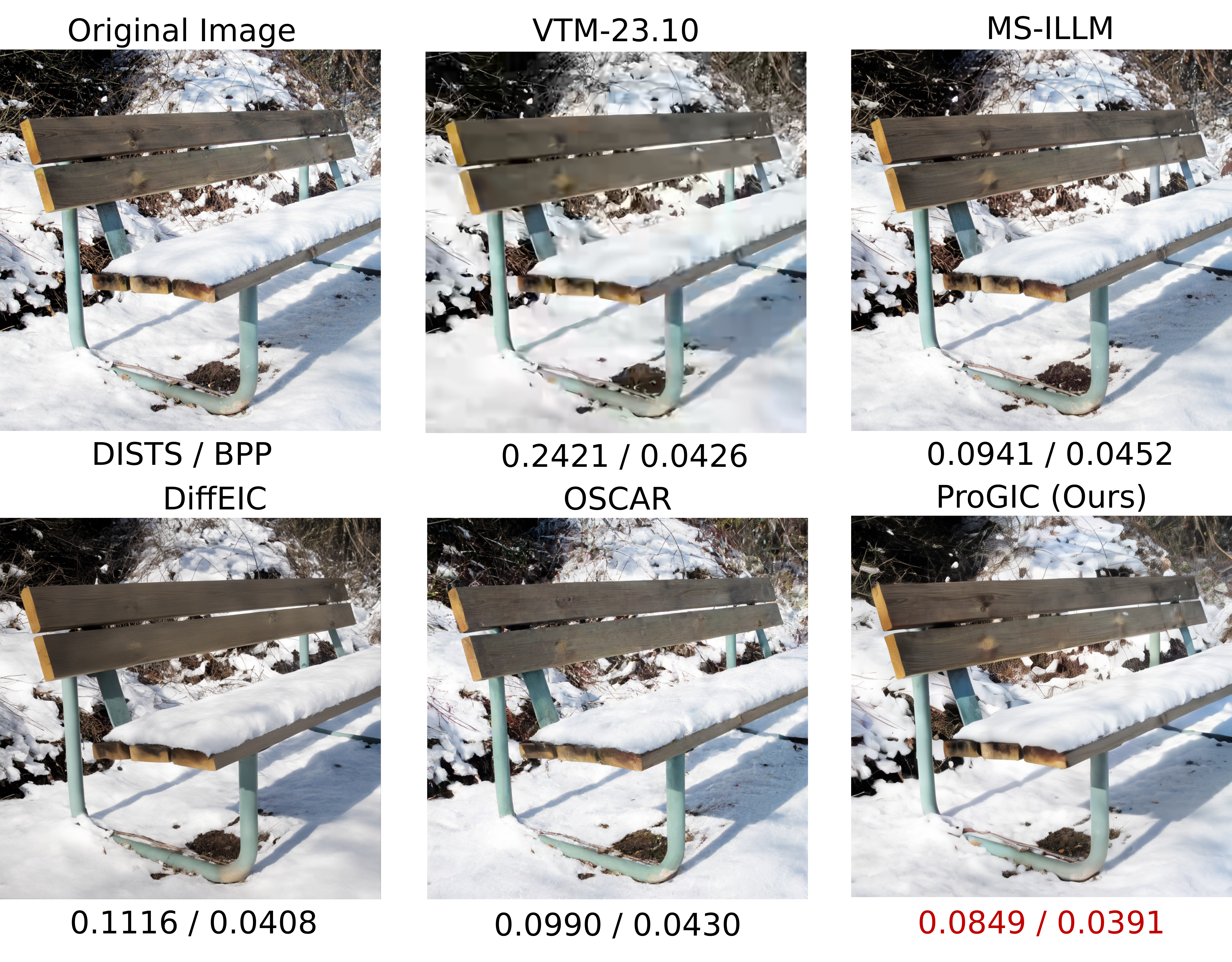}
  \caption{Visualization of reconstructed images from different methods on Tecnick. Values denote DISTS / BPP. Lower DISTS indicates better perceptual quality, and lower BPP indicates higher compression.}
  \label{fig:tecnick1}
\end{figure*}

\begin{figure*}[t]
  \centering
  \includegraphics[width=0.99\linewidth]{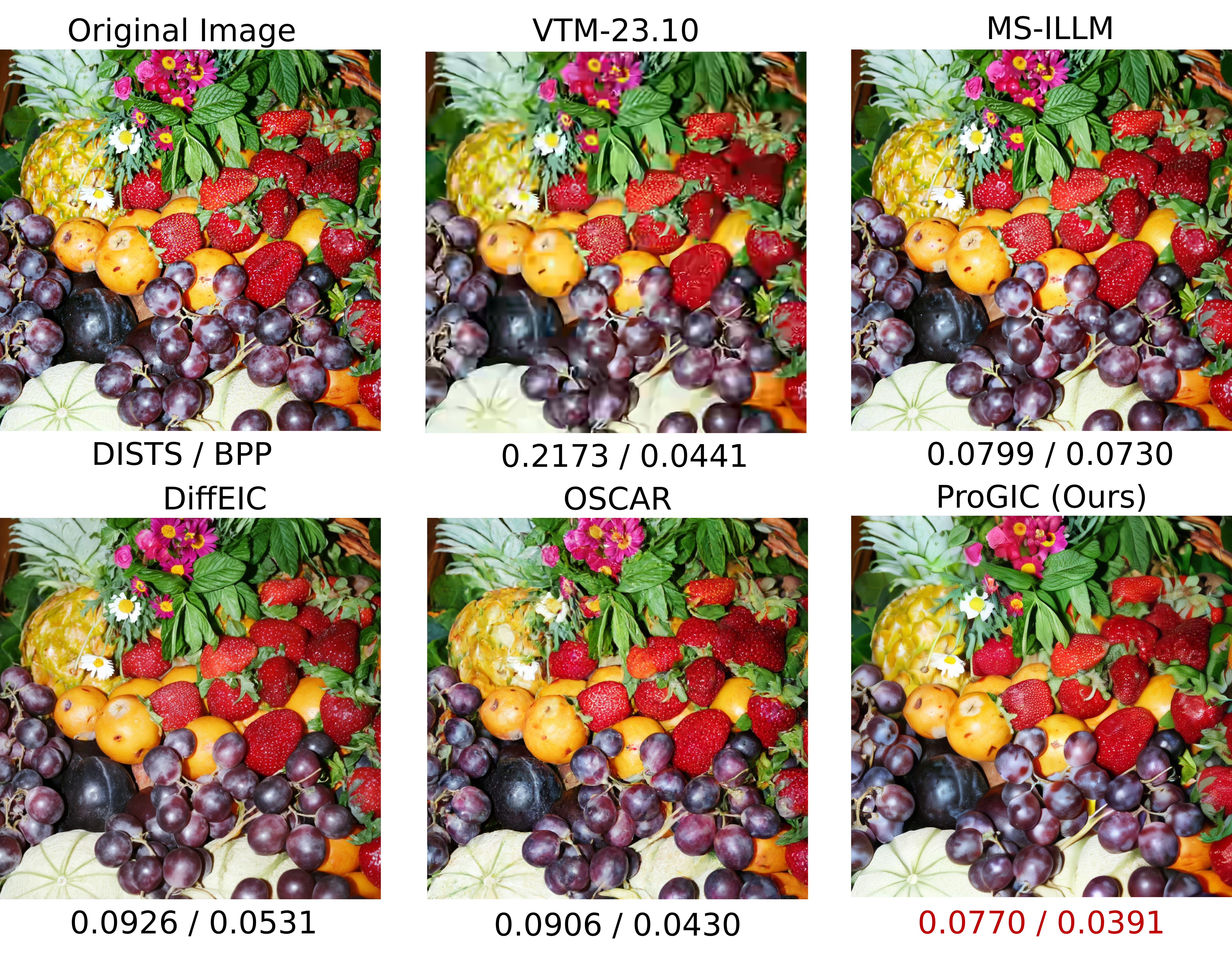}
  \caption{Visualization of reconstructed images from different methods on Tecnick. Values denote DISTS / BPP. Lower DISTS indicates better perceptual quality, and lower BPP indicates higher compression.}
  \label{fig:tecnick2}
\end{figure*}

\begin{figure*}[t]
  \centering
  \includegraphics[width=0.60\linewidth]{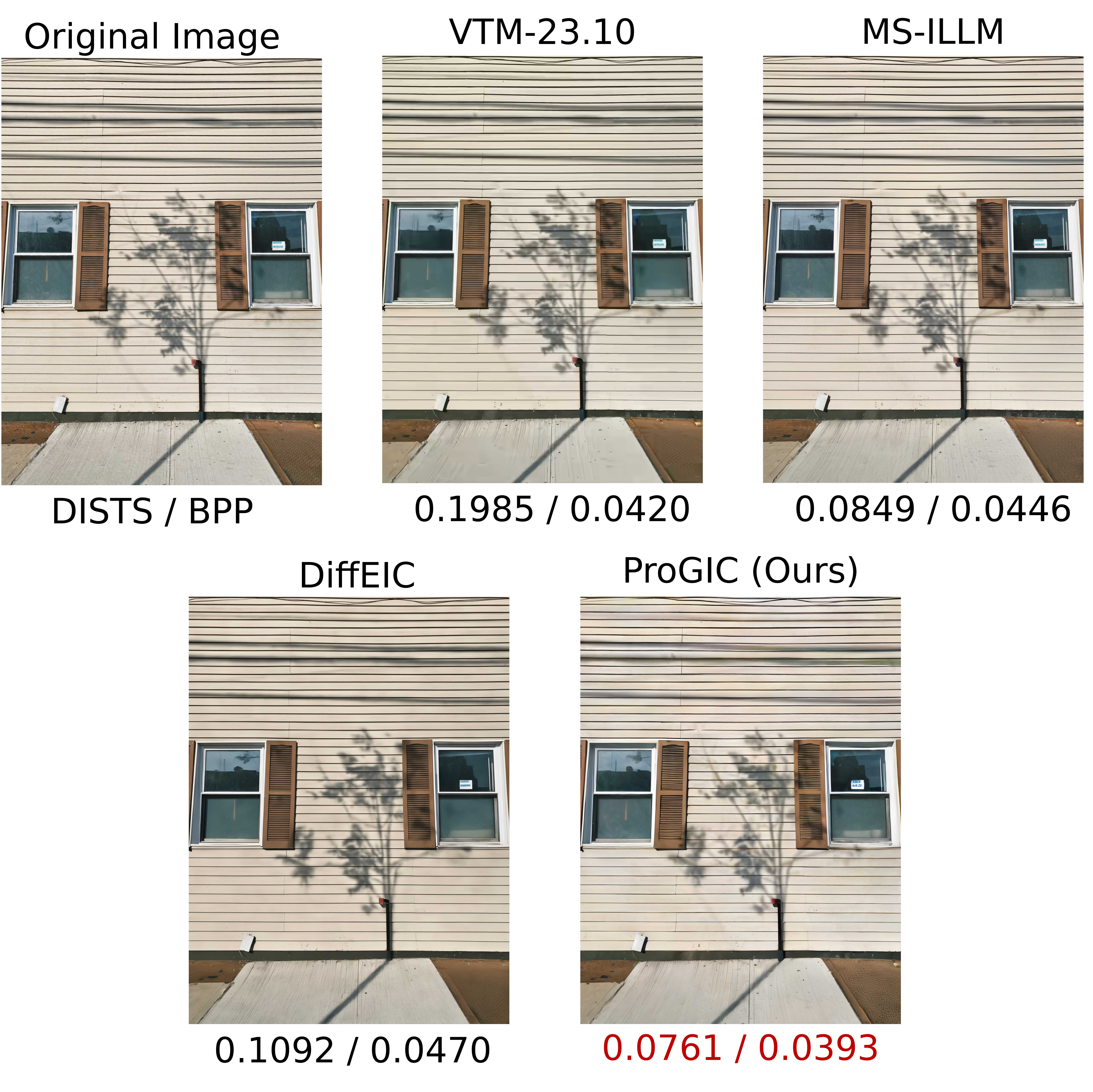}
  \caption{Visualization of reconstructed images from different methods on CLIC 2020. Values denote DISTS / BPP. Lower DISTS indicates better perceptual quality, and lower BPP indicates higher compression.}
  \label{fig:clic2}
\end{figure*}

In the main text, we present visual comparisons of ProGIC against other methods on the Kodak~\cite{Kodak} dataset. In this section, we provide additional qualitative results on the Tecnick~\cite{tecnick}, DIV2K~\cite{div2k}, and CLIC 2020~\cite{clic2020} datasets. The result are shown in~\cref{fig:tecnick1,fig:tecnick2,fig:div2k1,fig:div2k2,fig:clic1,fig:clic2}.

\begin{figure*}[t]
  \centering
  \includegraphics[width=0.99\linewidth]{fig/div2k1.pdf}
  \caption{Visualization of reconstructed images from different methods on DIV2K. Values denote DISTS / BPP. Lower DISTS indicates better perceptual quality, and lower BPP indicates higher compression.}
  \label{fig:div2k1}
\end{figure*}

\begin{figure*}[t]
  \centering
  \includegraphics[width=0.99\linewidth]{fig/div2k2.pdf}
  \caption{Visualization of reconstructed images from different methods on DIV2K. Values denote DISTS / BPP. Lower DISTS indicates better perceptual quality, and lower BPP indicates higher compression.}
  \label{fig:div2k2}
\end{figure*}

\begin{figure*}[t]
  \centering
  \includegraphics[width=0.90\linewidth]{fig/clic1.pdf}
  \caption{Visualization of reconstructed images from different methods on CLIC 2020. Values denote DISTS / BPP. Lower DISTS indicates better perceptual quality, and lower BPP indicates higher compression.}
  \label{fig:clic1}
\end{figure*}

\end{document}